\documentclass[journal]{IEEEtran}
\usepackage{times}
\usepackage{epsfig}
\usepackage{graphicx}
\usepackage{amsmath}
\usepackage{amssymb}
\usepackage{multirow}
\usepackage{mathrsfs}
\usepackage{setspace}
\usepackage{colortbl}
\definecolor{lightgray}{gray}{.92}
\definecolor{tinygray}{gray}{.96}

\usepackage[caption=false,font=footnotesize]{subfig}

\newcommand{\tabincell}[2]{\begin{tabular}{@{}#1@{}}#2\end{tabular}}

\usepackage{hyperref}

\ifCLASSINFOpdf
\else
\fi
\begin{document}
%
\title{Remove Cosine Window from Correlation Filter-based Visual Trackers: When and How}
\author{Feng~Li,
        Xiaohe~Wu, 
        Wangmeng~Zuo,
        David~Zhang,
        and~Lei~Zhang
\thanks{This work is partially supported by the National Natural Scientific Foundation of China (NSFC) under Grant No. 61671182 and 61872118, and the HK RGC GRF grant (under no. PolyU 152124/15E).}
\thanks{F. Li, X. Wu, and W. Zuo are with the School of Computer Science and Technology, Harbin Institute of Technology, Harbin, 150001, China. e-mail: (fengli$\_$hit@hotmail.com, xhwu.cpsl.hit@gmail.com, wmzuo@hit.edu.cn).}
\thanks{D. Zhang is with the School of Science and Engineering, The Chinese University of Hong Kong (Shenzhen), Shenzhen, China, e-mail: (csdzhang@comp.polyu.edu.hk).}
\thanks{L. Zhang is with the Department of Computing, The Hong Kong Polytechnic University, Hong Kong, e-mail: (davidzhang@cuhk.edu.cn).}
\thanks{(Corresponding author: Wangmeng Zuo)}
\thanks{Manuscript received May xx, 2019.}
}

\markboth{Journal of \LaTeX\ Class Files,~Vol.~14, No.~8, August~2015}%
{Shell \MakeLowercase{\textit{et al.}}: Bare Demo of IEEEtran.cls for IEEE Journals}

\maketitle

\begin{abstract}
   Correlation filters (CFs) have been continuously advancing the state-of-the-art tracking performance and have been extensively studied in the recent few years.
   Most of the existing CF trackers adopt a cosine window to spatially reweight base image to alleviate boundary discontinuity.
   However, cosine window emphasizes more on the central region of base image and has the risk of contaminating negative training samples during model learning.
   On the other hand, spatial regularization deployed in many recent CF trackers plays a similar role as cosine window by enforcing spatial penalty on CF coefficients.
   Therefore, we in this paper investigate the feasibility to remove cosine window from CF trackers with spatial regularization.
   When simply removing cosine window, CF with spatial regularization still suffers from small degree of boundary discontinuity.
   To tackle this issue, binary and Gaussian shaped mask functions are further introduced for eliminating boundary discontinuity while reweighting the estimation error of each training sample, and can be incorporated with multiple CF trackers with spatial regularization.
   In comparison to the counterparts with cosine window, our methods are effective in handling boundary discontinuity and sample contamination, thereby benefiting tracking performance.
   Extensive experiments on three benchmarks show that our methods perform favorably against the state-of-the-art trackers using either handcrafted or deep CNN features. The code is publicly available at \emph{\url{https://github.com/lifeng9472/Removing_cosine_window_from_CF_trackers}}.
\end{abstract}


\section{Introduction}

\IEEEPARstart{C}{orrelation} filter (CF) is a representative framework for visual tracking and has attracted great research interest.
Since the pioneering work of MOSSE~\cite{bolme2010visual}, extensive studies have been given to improve the CF models by incorporating nonlinear kernel~\cite{henriques2015high,tang2015multi}, scale adaptivity~\cite{danelljan2016discriminative,li2014scale,Li2017Integrating}, max-margin
classification~\cite{zuo2016learning},  spatial regularization~\cite{danelljan2015learning,GaloogahiSL14,Luke2016Discriminative}, and continuous convolution~\cite{Danelljan2016CCOT,Danelljan2016ECO}.
Moreover, the use of deep representation and its combination with handcrafted features also significantly boosts the tracking performance.
Benefited from the progress in models and feature representation, CFs have continuously advanced the state-of-the-art tracking accuracy and robustness in the recent few years.

\begin{figure*}[!htbp]
   \setlength{\belowcaptionskip}{-0.1cm}
   \centering
   \subfloat[KCF without cosine window]{
   \includegraphics[width=0.45\textwidth]{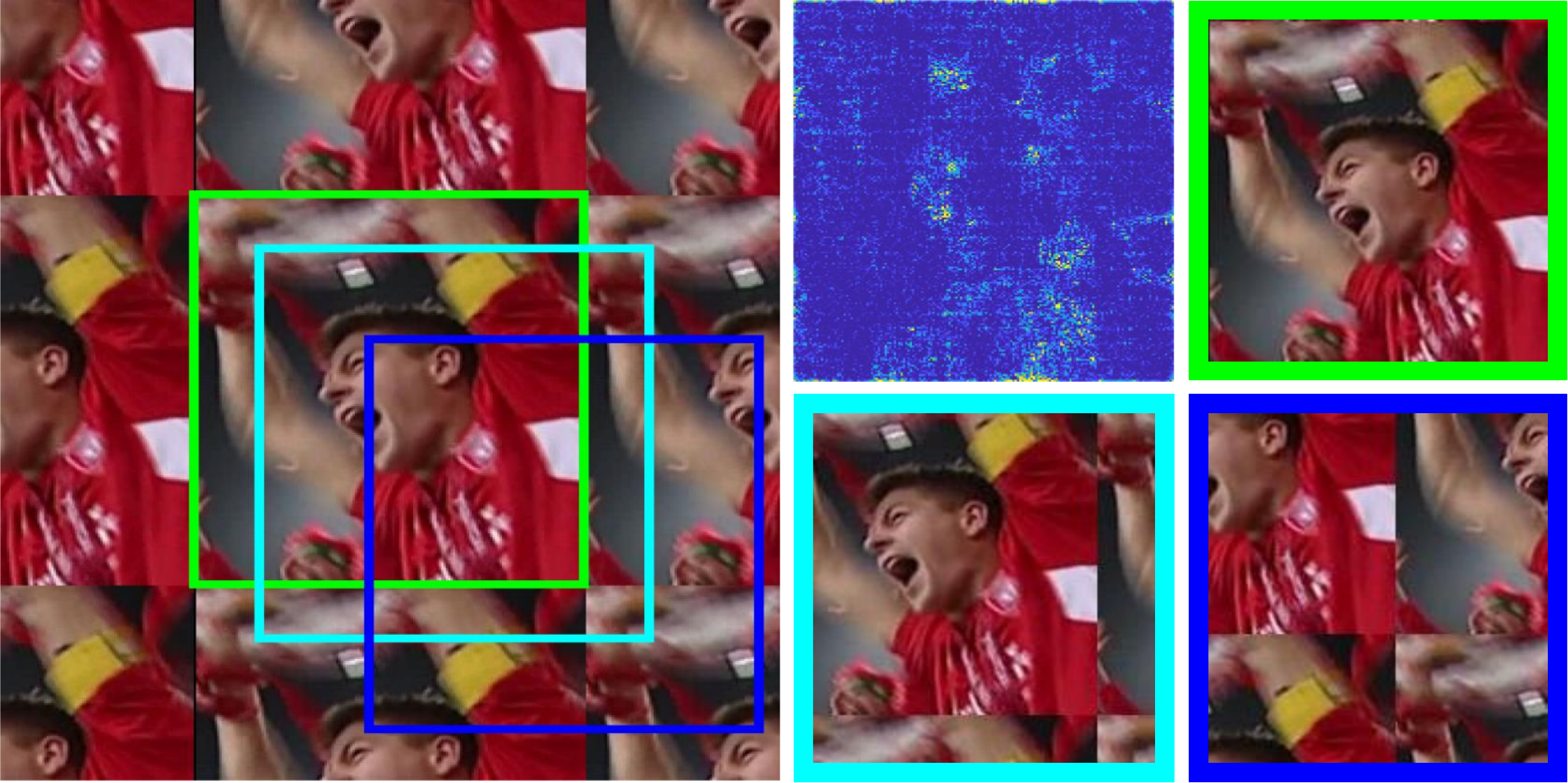}}
   \hspace{0.005\textwidth}
   \subfloat[KCF with cosine window]{
   \includegraphics[width=0.45\textwidth]{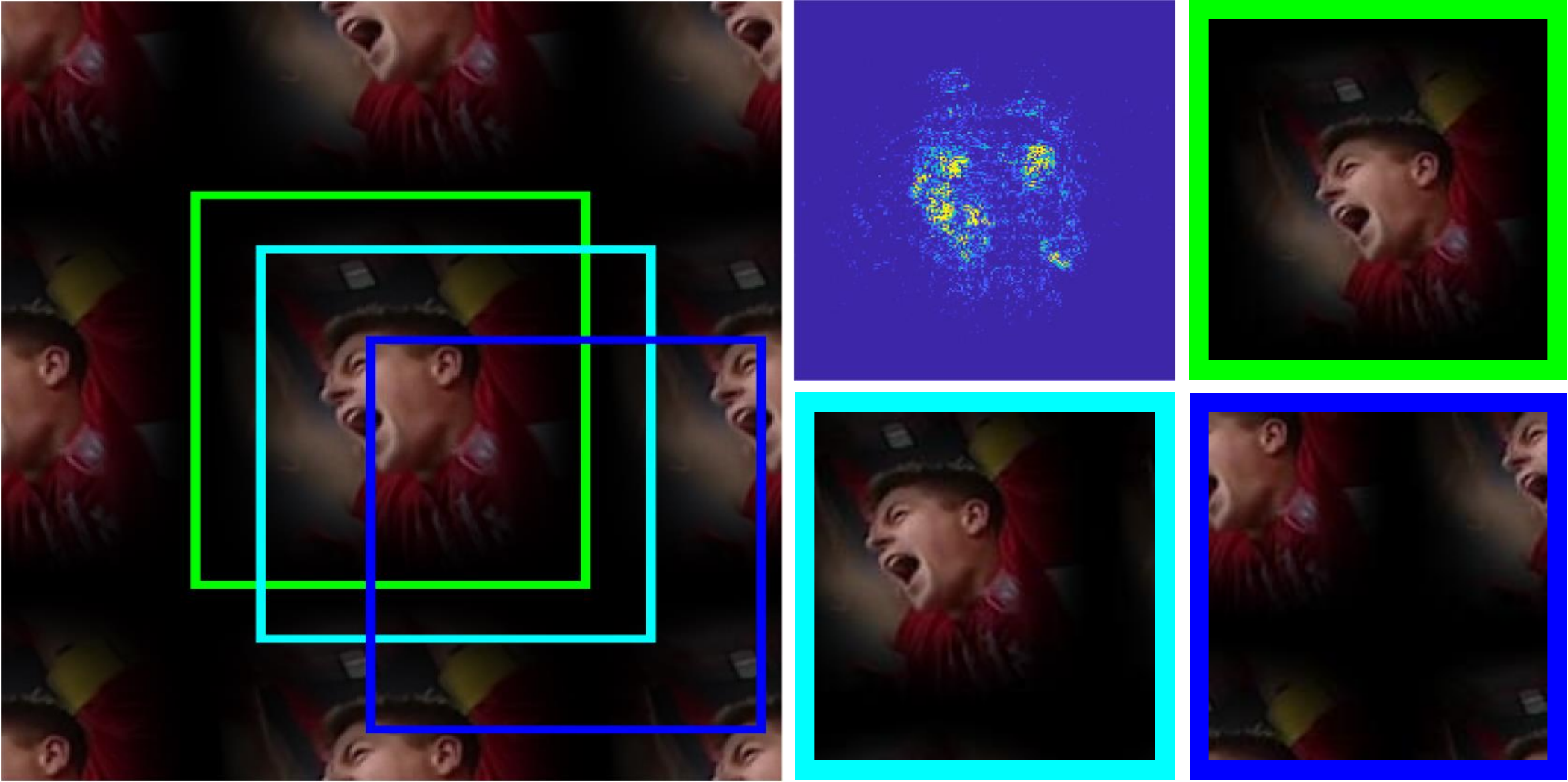}}\\
   \vspace{-8pt}
   \subfloat[BACF with cosine window]{
   \includegraphics[width=0.45\textwidth]{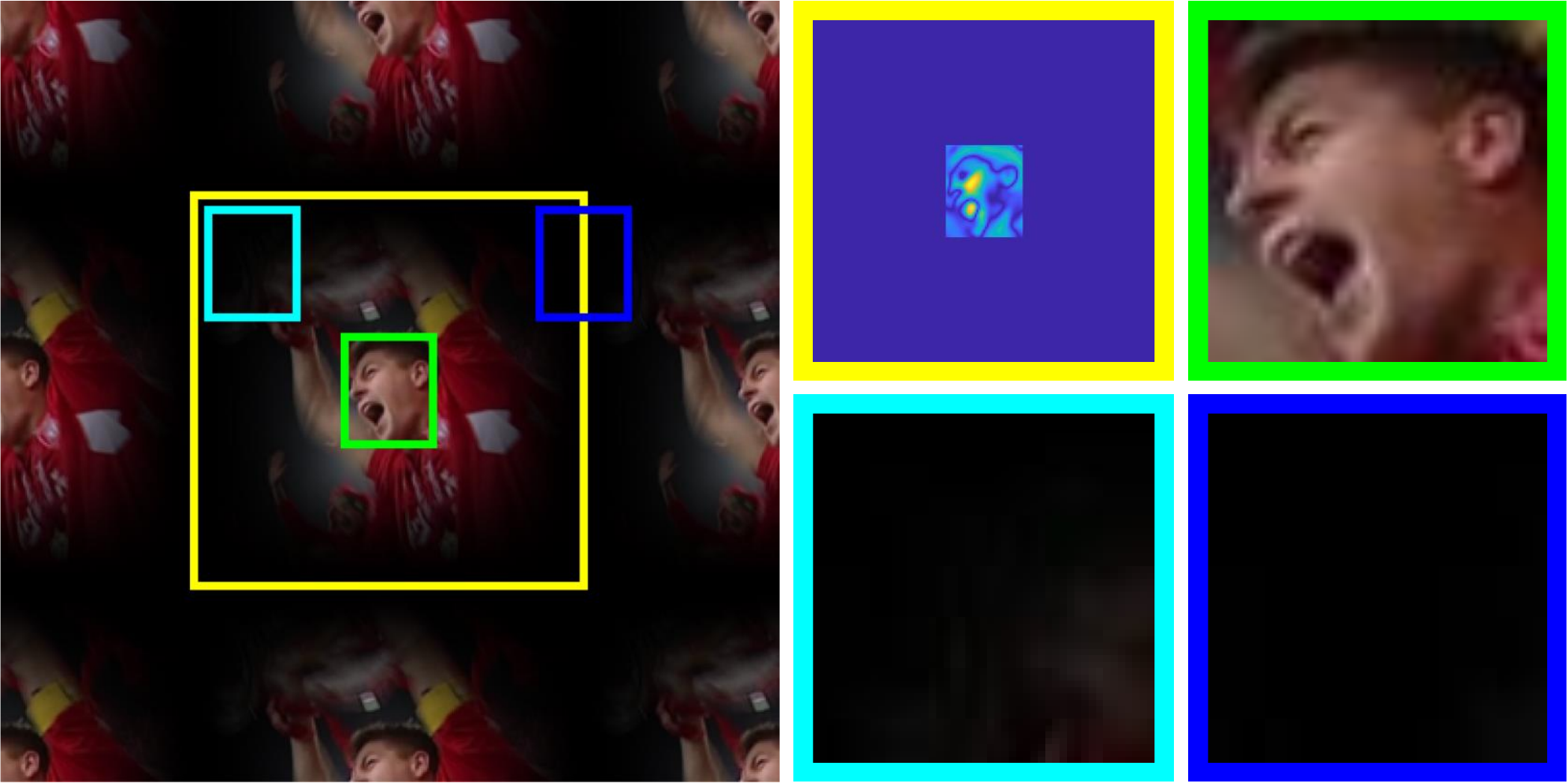}}
   \hspace{0.005\textwidth}
   \subfloat[BACF without cosine window]{
   \includegraphics[width=0.45\textwidth]{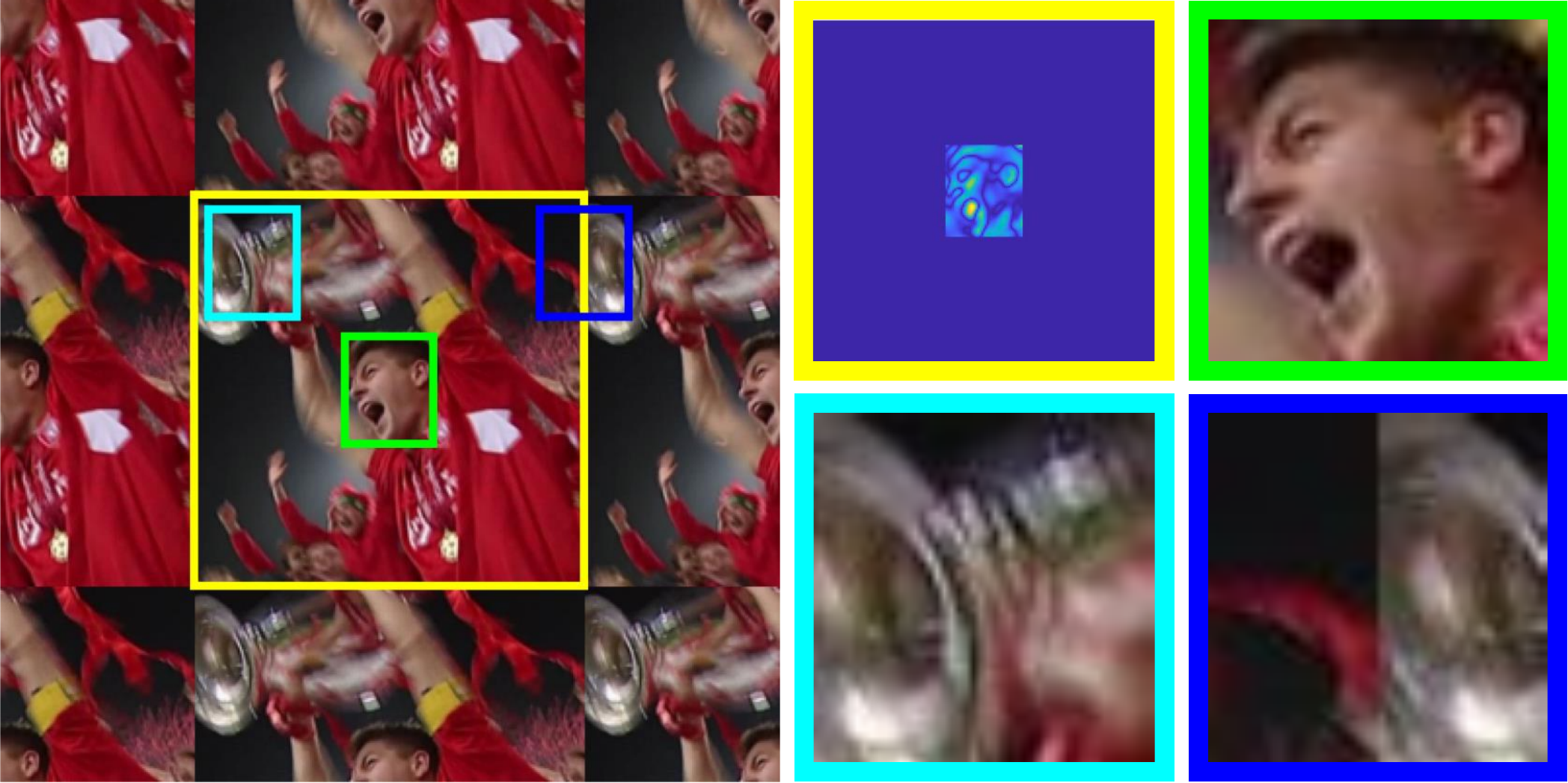}}
 \caption{Illustration of the cyclic extension of base image, representative training samples and learned filters by KCF and BACF with and without cosine window, including KCF (a) without and (b) with cosine window, BACF (c) with and (d) without cosine window. In each of (a)(b)(c)(d), the left part shows the cyclic extension of base image, while the right part illustrates the learned filter and three training samples cropped from the left part.
   }
   \label{fig:imfiltershow}
   \end{figure*}

In standard CF, the training set is formed as all the cyclic shifts of a base image and can be represented as a circulant matrix, making that CFs can be efficiently learned via fast Fourier transform (FFT).
Albeit such circulant property greatly benefits learning efficiency, the negative samples (i.e., shifted images) will suffer from the boundary discontinuity problem.
As shown in Fig.~\ref{fig:imfiltershow}(a), except for the base image in {\color{green}green} box, all the shifted images (e.g., the two patches in {\color{cyan}cyan} and {\color{blue}blue} boxes) are generated using the circulant property and are not truly negative patches in real images.
%


In order to alleviate boundary discontinuity, cosine window has been introduced in early CF trackers, e.g., MOSSE~\cite{bolme2010visual} and KCF~\cite{henriques2015high}, and generally inherited by the subsequent improved models~\cite{danelljan2016discriminative,danelljan2015learning,Galoogahi2017Learning}.
In particular, cosine window bands on base image as a pre-processing step by multiplying with a cosine shaped function (i.e., larger values for central regions and zeros for boundary pixels).
Using KCF~\cite{henriques2015high} as an example, it can be seen from Fig.~\ref{fig:imfiltershow}(b) that after deploying cosine window boundary discontinuity can be largely suppressed (e.g., the patch in {\color{cyan}cyan} box).
Nonetheless, the shifted images near boundary are still plagued, as shown in the patch in {\color{blue}blue} box.
Moreover, cosine window is deployed to base image, thus has the risk of contaminating negative training samples into unreal image patches.


Recently, spatial regularization has also been suggested in numerous CF trackers~\cite{danelljan2015learning, Danelljan2016ECO, GaloogahiSL14, Galoogahi2017Learning, li2018learning, bhat2018unveiling} to alleviate boundary discontinuity, which can be roughly grouped into two categories.
On the one hand, SRDCF~\cite{danelljan2015learning} and its later works~\cite{Danelljan2016ECO,li2018learning,bhat2018unveiling} penalize the filter coefficients near boundaries to approximate zero.
On the other hand, CFLB~\cite{GaloogahiSL14} and its multi-channel extension BACF~\cite{Galoogahi2017Learning} directly restrict the filter coefficients to be zero outside target bounding boxes.
In general, existing CF trackers with spatial regularization still adopt cosine window, and are more effective in handling boundary discontinuity, as illustrated in Fig.~\ref{fig:imfiltershow}(c).
Even though, the contamination of negative samples remains inevitable and may give rise to degraded performance.

Comparing the filters in Fig.~\ref{fig:imfiltershow}(a)(b)(c), one can see that cosine window plays a similar role as spatial regularization in enforcing the filter coefficients near boundary to approach zero.
Therefore, it is interesting to ask the first problem concerned in this work: when can we remove cosine window from CFs?
Using BACF as an example, Fig.~\ref{fig:imfiltershow}(d) shows the learned filters by simply removing cosine window.
It can be observed that the filters in Fig.~\ref{fig:imfiltershow}(c)(d) are only moderately different in appearance.
Our empirical study further shows that BACF without cosine window performs slightly inferior to BACF.
Thus, our answer to this question is when spatial regularization is deployed, it is possible to remove cosine window from CF trackers.

The second problem concerned in this work is: how to remove cosine window from CF trackers with spatial regularization.
To begin with, Fig.~\ref{fig:imfiltershow}(d) illustrates three representative samples used in BACF by simply removing cosine window.
While most samples are real image patches (e.g., those in {\color{green}green} and {\color{cyan}cyan} boxes), there remain a small percentage of negative samples suffering from boundary discontinuity (e.g., the patch near boundary in {\color{blue}blue} box).
To address this issue, we introduce a binary mask function to eliminate the effect of boundary discontinuous sample.
In particular, we assign zero to negative samples with discontinuous boundaries, thereby safely removing cosine window.
To further improve tracking performance, a Gaussian shaped mask function is also presented to emphasize more on samples near target center.



To evaluate the feasibility and effectiveness of removing cosine window, we incorporate our methods with several representative CF trackers with spatial regularization, including BACF~\cite{Galoogahi2017Learning}, STRCF~\cite{li2018learning}, ECO~\cite{Danelljan2016ECO}, and UPDT~\cite{bhat2018unveiling}.
Experiments are then conducted on three tracking benchmarks, i.e., OTB-2015~\cite{wu2015object}, Temple-Color~\cite{Liang2015Encoding} and VOT-2018~\cite{kristan2018sixth}.
In comparison to the counterparts with cosine window, our methods are effective in handling boundary discontinuity while avoiding sample contamination, and give rise to more robust appearance model as well as tracking performance.
Moreover, by incorporating with UPDT~\cite{bhat2018unveiling},
our methods achieve the state-of-the-art tracking performance, and attain an EAO score of 0.391 on VOT-2018, surpassing the rank-1 tracker (i.e., LADCF~\cite{xu2018learning}) in the VOT2018 challenge.

To sum up, the main contributions of this paper are:
\begin{itemize}
   \item \emph{When to remove cosine window from CF trackers?}
       Our analysis and empirical study show that both spatial regularization and cosine window can be used to alleviate boundary discontinuity.
       When spatial regularization is deployed, it is possible to remove cosine window from CF trackers.
       %
       %
   %
   \item \emph{How to remove cosine window from CF trackers with spatial regularization?}
       When removing cosine window from CF trackers with spatial regularization, there are still a small percentage of negative samples suffering from boundary discontinuity.
       To tackle this issue, two mask functions are introduced for reweighting the estimation error of each training sample.
       And our methods can be incorporated with multiple representative CF trackers with spatial regularization.
   \item Experiments on three tracking benchmarks indicate that our methods can eliminate boundary discontinuity while avoiding sample contamination, and perform favorably against the state-of-the-art trackers.
   %
\end{itemize}

The remainder of this paper is organized as follows.
Section~\ref{sec:RelatedWork} briefly reviews the CF trackers relevant to this work.
Section~\ref{sec:when} provides both qualitative and quantitative analysis to dissect the effect of removing cosine window from CF trackers.
Section~\ref{sec:howto} further describes our solutions to remove cosine window, which are then incorporated with multiple CF trackers with spatial regularization.
Section~\ref{sec:exp} reports the experimental results.
Finally, Section~\ref{sec:conclusion} ends this work with several concluding remarks.

\section{Related Work}\label{sec:RelatedWork}

%






The core problem of CF trackers is to learn a discriminative filter for the next frame from current frame and historical information.
Early methods, e.g., MOSSE~\cite{bolme2010visual} and KCF~\cite{henriques2015high}, formulate the CF framework with one single base image from the current frame, and update the CFs using the linear interpolation strategy.
Denote by the sample pair $\{\left(\mathbf{x}_t,\mathbf{y}_t\right)\}$ in frame $t$, where each sample $\mathbf{x}_t$ consists of $L$ feature maps with $\mathbf{x}_t = [\mathbf{x}_{t,1}, ..., \mathbf{x}_{t,L}]$, and $\mathbf{y}_t$ represents the Gaussian shaped label.
Then the correlation filter $\mathbf{f}$ is obtained by minimizing the following objective,
\begin{equation}\label{equ:CF_single}
\small
\mathcal{E}\left ( \mathbf{f} \right ) = \frac{1}{2}\left \| \sum_{l=1}^L \mathbf{f}_l \star \left(\mathbf{x}_{t,l} \odot \mathbf{c}\right)- \mathbf{y}_t \right \|^2 + {\lambda} \mathcal{R} ( \mathbf{f} ),
\end{equation}
where $\star$ and $\odot$ respectively stand for circular convolution and Hadamard product, $\mathbf{c}$ denotes cosine window, and $\lambda$ denotes the tradeoff parameter of the regularization term $\mathcal{R}(\mathbf{f})$.

Since the pioneering work of MOSSE~\cite{bolme2010visual}, many improvements have been made to CF trackers.
On the one hand, the CF models have been consistently improved with the introduction of non-linear kernel~\cite{henriques2015high}, scale adaptivity~\cite{danelljan2016discriminative,li2014scale,Li2017Integrating},
long-term tracking~\cite{Ma2015Long}, part-based CFs~\cite{RealTimeCF2018}, particle filters~\cite{ParticleCF2018}, spatial regularization~\cite{GaloogahiSL14,li2018learning}, continuous convolution~\cite{Danelljan2016CCOT,Danelljan2016ECO}, and formulation with multiple base images~\cite{danelljan2015learning,Danelljan2016ECO,bhat2018unveiling}.
On the other hand, progress in feature engineering, e.g., HOG~\cite{DalalHOG2005}, ColorName~\cite{Danelljan2014Adaptive} and deep CNN features~\cite{ma2015hierarchical,qi2016hedged,gladh2016deep}, also greatly benefits the performance of CF trackers.

%

Among these improvements, we specifically mention a category of CF formulations with multiple base images~\cite{danelljan2015learning,Danelljan2016ECO,bhat2018unveiling}.
Given a set of $K$ base images $\{\left(\mathbf{x}_k,\mathbf{y}_k\right)\}_{k=1}^K$, CF with multiple base images can then be expressed as,
\begin{equation}\label{equ:CF}
   \small
\mathcal{E}\left ( \mathbf{f} \right ) = \frac{1}{2}\sum_{k=1}^K \alpha_k\left \| \sum_{l=1}^L \mathbf{f}_l \star \left(\mathbf{x}_{k,l} \odot \mathbf{c}\right)- \mathbf{y}_k \right \|^2 + {\lambda} \mathcal{R} ( \mathbf{f} ),
\end{equation}
where $\alpha_k$ represents the weight of the $k$-th base image $\mathbf{x}_k$.
For example, SRDCF~\cite{danelljan2015learning} and CCOT~\cite{Danelljan2016CCOT} simply adopt the latest $K$ frames as base images.
In SRDCFdecon~\cite{Danelljan2016Adaptive}, an adaptive decontamination model is presented to downweight corrupted samples while up-weighting faithful ones.
ECO~\cite{Danelljan2016ECO} and UPDT~\cite{bhat2018unveiling} apply a Gaussian mixture model (GMM) to determine both the weights as well as base images.
In general, CF trackers with multiple base images perform much better than those with single base image, and have achieved state-of-the-art tracking performance.


In contrast to CF with single base image in Eqn.~(\ref{equ:CF_single}), the introduction of multiple base images breaks the circulant structure, and generally requires iterative optimization algorithms to solve the resulting formulation in Eqn.~(\ref{equ:CF}).
Therefore, in this work different solutions are respectively developed for removing cosine window from CF trackers with single and multiple base images.

   \begin{table}[!htb]
      \renewcommand\arraystretch{1.35}
      \caption{The EAO, accuracy and robustness of two CF trackers (i.e., KCF and BACF) and their counterparts without using cosine window during training (i.e., KCF$_{RC}$ and BACF$_{RC}$) on the VOT-2018 dataset. Here, $\uparrow$ ($\downarrow$) denotes higher (lower) is better.
      }
      \scalebox{0.87}{
      \begin{tabular}{|c||c|c|c|c|}
      \hline
      \rowcolor{lightgray}
       \textbf{Methods} & \textbf{KCF}~\cite{henriques2015high}  &\textbf{KCF}$_{RC}$&\textbf{BACF}~\cite{Galoogahi2017Learning}  & \textbf{BACF}$_{RC}$ \\ \hline\hline
      \textbf{EAO ($\uparrow$)}        & 0.106                              & 0.069     &  0.137   & 0.124   \\
      \rowcolor{tinygray}
      \textbf{Accuracy ($\uparrow$)}   & 0.327                               & 0.374   &  0.432   &  0.466 \\
      \textbf{Robustness ($\downarrow$)} & 1.182                            &   1.823     & 0.757   & 0.892 \\ \hline
      \end{tabular}}
      \label{Tab:KCFandBACF}
   \end{table}

   %
   %

   \section{When to Remove Cosine Window}\label{sec:when}

   Cosine window is first introduced in the early MOSSE and KCF methods to alleviate the effect of boundary discontinuity, and then adopted in all the subsequent CF trackers.
   In the recent few years, spatial regularization has also been deployed in CF trackers for handling boundary discontinuity.
   Albeit cosine window is also adopted in CF with spatial regularization, considering their similar roles, it is natural to ask whether it is possible to remove cosine window from CF when spatial regularization is adopted.

   In this section, we use KCF and BACF as two representative examples, and evaluate the performance of CF trackers with and without cosine window on the VOT-2018 dataset.
   Here we name KCF and BACF without cosine window as KCF$_{RC}$ and BACF$_{RC}$, respectively.
   Table~\ref{Tab:KCFandBACF} lists their EAO, accuracy and robustness on VOT-2018.
   For KCF, it can be seen that removing cosine window is harmful to tracking performance and gives rise to an obvious EAO drop from 0.106 to 0.069.
   From Fig.~\ref{fig:imfiltershow}(a)(b), the filter learned by KCF is much different from that learned by KCF$_{RC}$ in appearance.
   Moreover, cosine window also performs similarly in enforcing non-central filter coefficients to approach zero.
   In contrast to KCF, cosine window actually plays a minor role on improving tracking performance for BACF, and the EAO of BACF$_{RC}$ is only 0.013 lower than that of BACF.
   From Fig.~\ref{fig:imfiltershow}(c)(d), the filters learned by BACF and BACF$_{RC}$ are also similar in appearance.
   Similar results can also be observed for STRCF~\cite{li2018learning}, ECO~\cite{Danelljan2016ECO} and UPDT~\cite{bhat2018unveiling} and on the OTB-2015 and Temple-Color benchmarks in Section \ref{sec:exp}, indicating that it is possible to remove cosine window from CF trackers when spatial regularization is introduced.

   \begin{figure*}[!htbp]
      \setlength{\belowcaptionskip}{0cm}
      \centering
      \subfloat[]{\label{fig:BACFSliding}
        \includegraphics[height = 4.9cm, width=4.9cm]{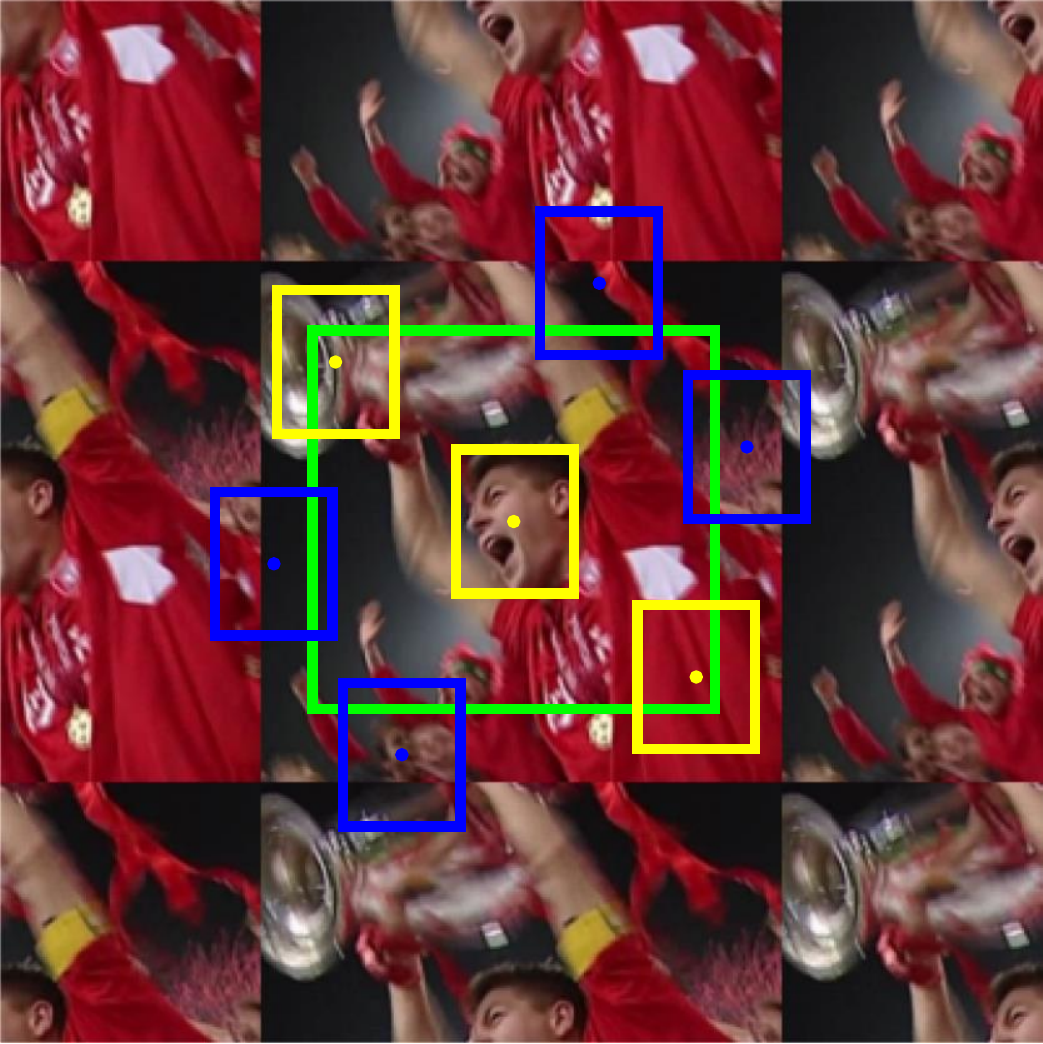}}
        \hspace{0.032\textwidth}
      \subfloat[]{\label{fig:MBinary}
        \includegraphics[height = 4.95cm, width=5.7cm]{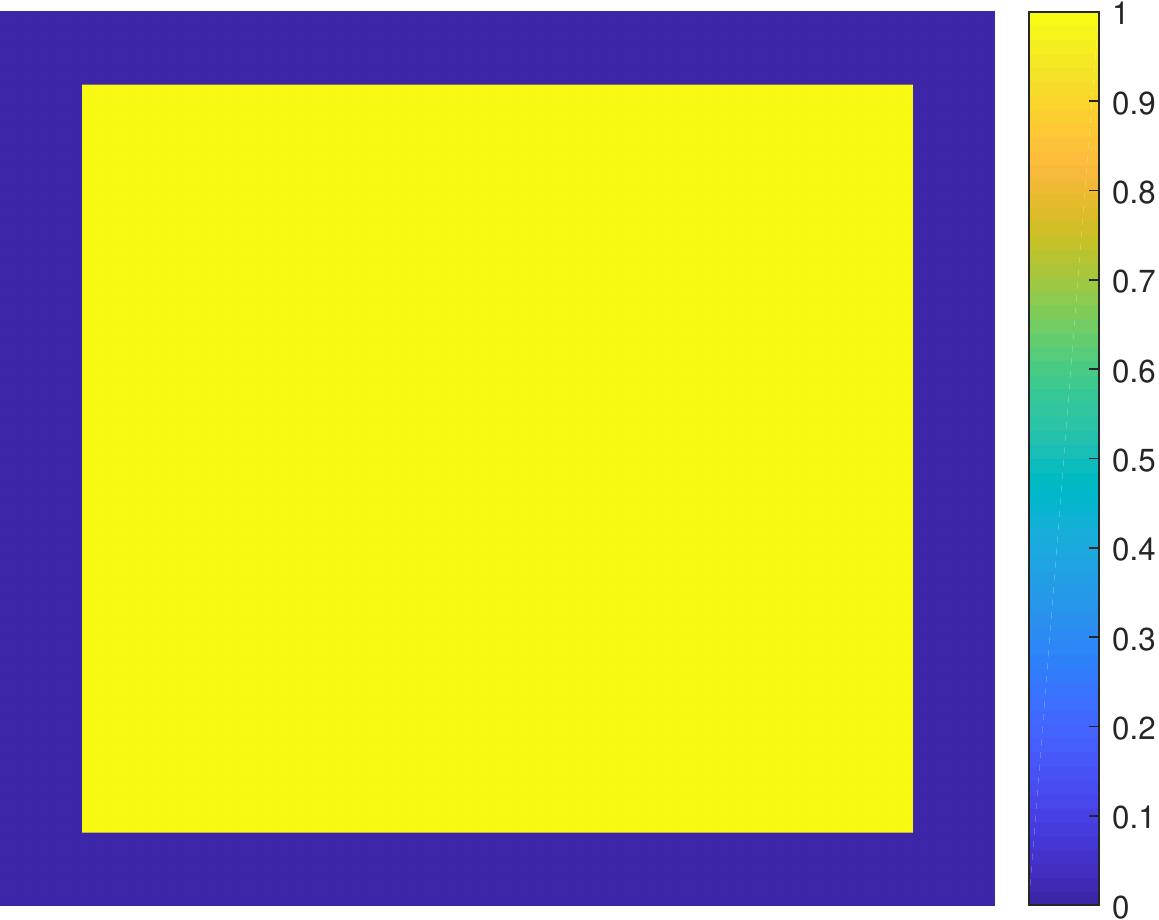}}
        \hspace{0.032\textwidth}
        \subfloat[]{\label{fig:MGaussian}
        \includegraphics[height = 4.95cm, width=5.7cm]{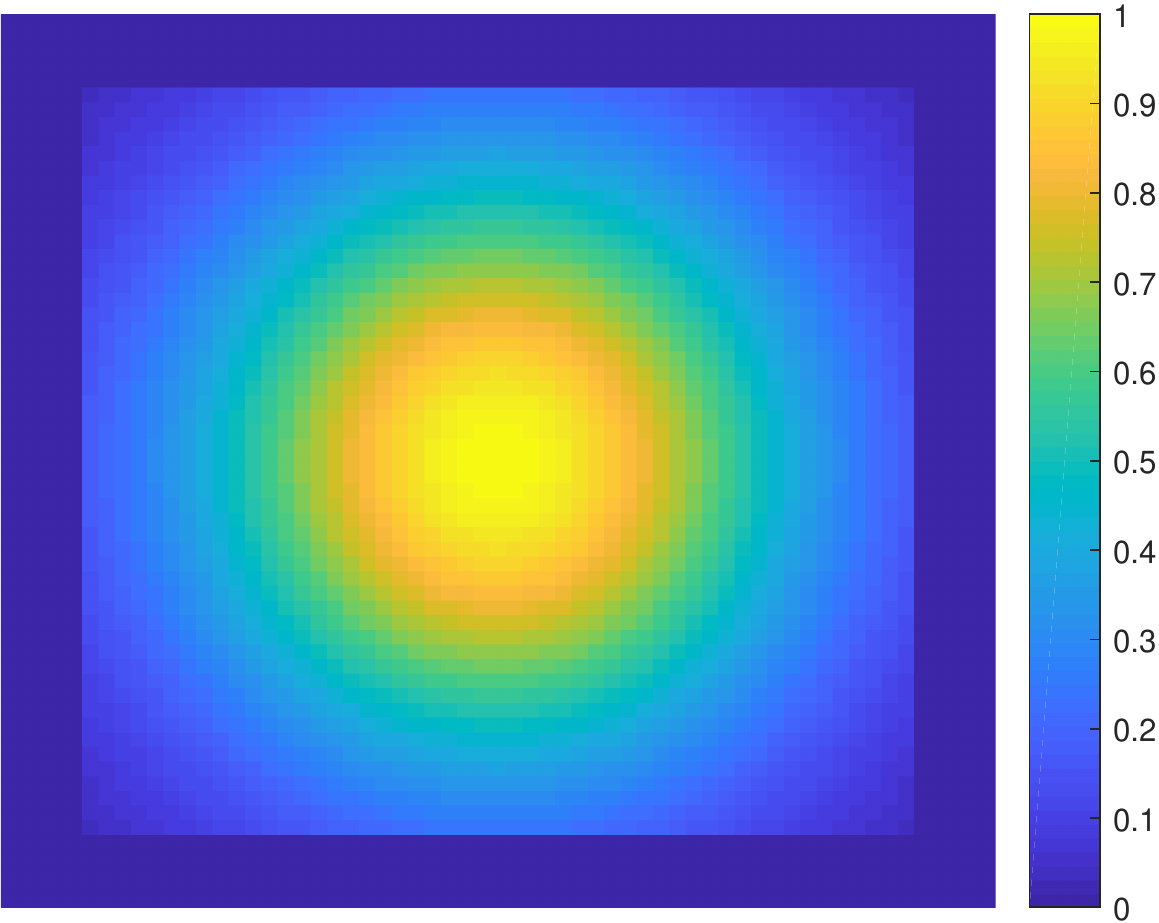}}
      \caption{Illustration of the mask functions. (a) Training samples from  on the video $\emph{soccer}$ used in BACF, where the samples outside {\color{green}green} box suffer from the boundary discontinuity problem.
      (b) The binary mask function $\mathbf{M}$. (c) The Gaussian shaped mask function $\mathbf{M}_G$.}
      \label{fig:Mmatrix}
      \end{figure*}

   We also note that BACF still slightly outperforms BACF$_{RC}$, which can be explained by taking both boundary discontinuity and sample contamination into account.
   From Fig.~\ref{fig:imfiltershow}(c), it can be seen that BACF can well handle boundary discontinuity by incorporating cosine window and spatial regularization.
   However, cosine window is deployed on the base image, which makes the shifted negative samples contaminated and may be harmful to tracking performance.
   In BACF$_{RC}$, one can see from Fig.~\ref{fig:imfiltershow}(d) that most negative samples are real image patches (e.g., those in {\color{green}green} and {\color{cyan}cyan} boxes).
   However, a small percentage of negative samples still suffer from boundary discontinuity (e.g., that in {\color{blue}blue} box), which may explain the slight inferiority of BACF$_{RC}$ in comparison to BACF.
   To sum up, for removing cosine window, it is better to avoid sample contamination as well as eliminate boundary discontinuity for all negative samples.
   Thus, we turn to the second problem of this work, i.e., how to remove cosine window from CF trackers with spatial regularization, and present our solutions in the next section.


   \section{How to Remove Cosine Window}\label{sec:howto}

   Simply removing cosine window from CF trackers with spatial regularization generally cannot outperform its counterpart due to that the negative samples near boundary still suffer from boundary discontinuity.
   To address this issue, we modify the formulation of CF trackers by introducing mask function to deactivate the boundary discontinuous samples.
   Two mask functions are presented to eliminate boundary discontinuity as well as emphasize more on samples near the target center.
   Then, optimization algorithms are respectively developed for removing cosine window from CF trackers with single and multiple base images.


   \subsection{Problem formulation}

   Without loss of generality, we use BACF as an example to analyze the positions of negative samples suffering from boundary discontinuity.
   Suppose that the sizes of the target bounding box and base image are $h \times w$ and $H \times W$, respectively.
   For BACF, we have $H = W = 5\sqrt{hw}$.
   From Fig.~\ref{fig:Mmatrix}(a), it can be seen that only the samples at position $(x, y)$ are with discontinuous boundaries when $ \frac{H}{2} \geq |x| > \frac{H}{2} - \frac{h}{2} $ and $ \frac{W}{2} \geq |y| > \frac{W}{2} - \frac{w}{2} $.
   In general, $H$ ($W$) is much larger than $h$ ($w$), and thus the majority of samples (e.g., $64\%$ when $h=w$) are real image patches.
%
%
%
%
%
%
%
   In order to eliminate the effect of boundary discontinuity, we introduce a binary mask function $\mathbf{M}$ shown in Fig.~\ref{fig:Mmatrix}(b) to indicate the samples of real image patches.
   In particular, a sample at position $(x, y)$ is a real image patch when ${M}(x, y) = 1$.
   Then the binary mask $\mathbf{M}$ can be defined as follows,
   \begin{equation}\label{equ:MDef}
      {M}\left ( x,y \right ) = \begin{cases}
         1, & \!\!\!\!\text{ if }  |x| \leq \frac{H}{2} - \frac{h}{2} \text{ and}\,\,  |y| \leq \frac{W}{2} - \frac{w}{2}, \\
         0, & \!\!\!\! \text{ otherwise.}
         \end{cases}
   \end{equation}


   In the following, we use Eqn.~(\ref{equ:CF}) as a general form to illustrate how to eliminate boundary discontinuity while avoiding sample contamination for CF trackers with spatial regularization.
   In particular, we remove cosine window from Eqn.~(\ref{equ:CF}), and incorporate the binary mask $\mathbf{M}$ to deactivate the negative samples suffering from boundary discontinuity, resulting in the following model,
   \begin{equation}\label{equ:MSRCF}
   \mathcal{E}\left ( \mathbf{f} \right ) = \frac{1}{2}\!\sum_{k=1}^K\! \alpha_k\!\left \| \mathbf{M} \odot \left ( \sum_{l=1}^L \mathbf{f}_l \star \mathbf{x}_{k,l} \!-\! \mathbf{y}_k \!\!\right ) \right \|^2 \!\!\!+\! {\lambda} \mathcal{R}\!\left ( \mathbf{f} \right ).
   \end{equation}
   With the introduction of $\mathbf{M}$, the estimation error of the sample with discontinuous boundary can be safely excluded during training.
   In comparison to CF tracker with cosine window in Eqn.~(\ref{equ:CF}), the formulation in Eqn.~(\ref{equ:MSRCF}) can circumvent both boundary discontinuity and sample contamination, thereby benefiting tracking performance.


   Furthermore, the CF model usually is learned from an unbalanced set containing few positive samples and a large amount of negative samples.
   The binary mask $\mathbf{M}$ treats all boundary continuous samples equally, and has the risk of degrading tracking performance due to vast negative samples.
   Considering that the samples near the target center are more important than those on image boundaries, we also present a Gaussian shaped mask function $\mathbf{M}_G$ defined as,
   \begin{equation}\label{equ:MDef2}
      {M}_G\!\left ( x,y \right ) \!\!=\!\! \begin{cases}
         e^{- (\frac{x}{h \delta})^2 \! - (\frac{y}{w\delta})^2 }, & \!\!\!\!\!\!\text{ if }  |x| \!\leq\! \frac{H}{2} \!-\! \frac{h}{2} \text{ and } |y| \!\leq\! \frac{W}{2} \!-\! \frac{w}{2}, \\
         0, & \!\!\!\!\!\!\text{ otherwise},
         \end{cases}
   \end{equation}
   where the parameter $\delta$ is introduced to control the weight decay speed of training samples.
   Empirical study also validates that Gaussian shaped mask function $\mathbf{M}_G$ generally performs moderately better than binary mask function $\mathbf{M}$ for CF trackers with spatial regularization.

   Given a specific CF tracker, we denote the models by (i) removing cosine window, (ii) removing cosine window and incorporating binary mask function, (iii) removing cosine window and incorporating Gaussian mask function as CF$_{RC}$, CF$_{RCB}$, and CF$_{RCG}$, respectively.
   In the following, we present the optimization algorithms to solve the model in Eqn.~(\ref{equ:MSRCF}) for CF trackers with single and multiple base images, respectively.

\subsection{Solution for CF trackers with single base image}

For CFLB~\cite{Galoogahi2014Multi}, BACF~\cite{Galoogahi2017Learning}, CSR-DCF~\cite{Luke2016Discriminative} and STRCF~\cite{li2018learning}, the filter is updated by solving some CF models defined on a single base image (i.e., the current frame).  %
In this case, the resulting constrained optimization problem can be efficiently solved via alternating minimization, in which each subproblem has the closed-form solution.
When removing cosine window from this category of CF trackers with spatial regularization, we rewrite the model in Eqn.~(\ref{equ:MSRCF}) as
\begin{equation}\label{equ:SSRCF}
\small
\mathcal{E}\left ( \mathbf{f} \right ) = \frac{1}{2}\! \left \| \mathbf{M} \odot \left ( \sum_{l=1}^L \mathbf{f}_l \star \mathbf{x}_{l} \!-\! \mathbf{y} \!\!\right ) \right \|^2 \!\!\!+\! {\lambda} \mathcal{R}\!\left ( \mathbf{f} \right ).
\end{equation}
Then, alternating minimization algorithms can also be extended to solve it.
In the following, we take BACF as an example, and present an alternating direction method of multipliers (ADMM) to optimize the resulting formulation.


With simple algebra, the original formulation of BACF can be equivalently rewritten as,
\begin{equation}\label{equ:BACF2}
   \small
   \begin{split}
   \mathcal{L}(\mathbf{g})  = \frac{1}{2}\left \| \sum_{l=1}^L \left(\mathbf{x}_{l} \odot \mathbf{c} \right) \star (\mathbf{P}^{\text{T}} \mathbf{g}_{l}) - \mathbf{y} \right \|^2 + \frac{\lambda}{2} \left \| \mathbf{g} \right \|^2,
   \end{split}
\end{equation}
{where $\mathbf{P}$ stands for the binary mask matrix which crops the central $D$ elements of $\mathbf{g}_l$ with the size of $T$.}
%
%
%
After removing cosine window and incorporating with the mask function $\mathbf{M}$, we further let $\mathbf{f}_l = \mathbf{P}^{\text{T}} \mathbf{g}_l$, and the modified BACF model can be formulated as,
\begin{equation}\label{equ:MBACF}
   \small
   \begin{split}
   \mathcal{L}(\mathbf{f}, \mathbf{g})  &= \frac{1}{2}\left \| \mathbf{M} \odot (\sum_{l=1}^L \mathbf{x}_{l} \star \mathbf{f}_{l} - \mathbf{y}) \right \|^2 + \frac{\lambda}{2} \left \| \mathbf{g} \right \|^2, \\
   &\text{s.t.}  \quad \mathbf{f}_l = \mathbf{P}^{\text{T}}\mathbf{g}_l.
   \end{split}
\end{equation}

The model in Eqn.~(\ref{equ:MBACF}) is still a convex optimization problem, can be solved with the ADMM algorithm.
To begin with, we introduce another auxiliary variable $\mathbf{z} = \sum_{l=1}^L \mathbf{x}_{l} \star \mathbf{f}_{l} - \mathbf{y}$, 
and reformulate Eqn.~(\ref{equ:MBACF}) as,
\begin{equation}\label{equ:MBACF2}
   \small
   \begin{split}
   \mathcal{L}(\mathbf{f},\mathbf{g},\mathbf{z})  &= \frac{1}{2}\left \| \mathbf{M} \odot \mathbf{z} \right \|^2 + \frac{\lambda}{2} \left \| \mathbf{g} \right \|^2, \\
   &\text{s.t.}  \quad \mathbf{f}_l = \mathbf{P}^{\text{T}} \mathbf{g}_l, \quad \mathbf{z} = \sum_{l=1}^L \mathbf{x}_{l} \star \mathbf{f}_l - \mathbf{y}.
   \end{split}
\end{equation}
Then the augmented Lagrangian function of Eqn.~(\ref{equ:MBACF2}) can be expressed as,
\begin{equation}\label{equ:MBACFLag}
   \small
   \begin{split}
      \mathcal{L}(\mathbf{f},\mathbf{g},\mathbf{z}, &\boldsymbol{\zeta},\boldsymbol{\gamma})\!= \frac{1}{2}\left \| \mathbf{M} \odot \mathbf{z} \right \|^2 + \frac{\lambda}{2} \left \| \mathbf{g} \right \|^2 \\
       &+ \sum_{l=1}^L \boldsymbol{\zeta}_l ^{\text{T}}(\mathbf{f}_l \!-\! \mathbf{P}^{\text{T}}\mathbf{g}_l) + \frac{\mu}{2}\sum_{l=1}^L \left \| \mathbf{f}_l\! -\! \mathbf{P}^{\text{T}}\mathbf{g}_l \right \|^2 \\
       &+\!\boldsymbol{\gamma}^{\text{T}}\!\!\left ( \sum_{l=1}^L \mathbf{x}_{l} \! \star \!\mathbf{f}_l \!-\! \mathbf{y}\!\! -\! \mathbf{z}\! \right )\!\! + \!\frac{\tau}{2}\left \| \sum_{l=1}^L \!\mathbf{x}_{l} \! \star \!\mathbf{f}_l\! -\! \mathbf{y}\!\! -\! \mathbf{z} \right \|^2\!,
   \end{split}
\end{equation}
where $\boldsymbol{\zeta}$, $\boldsymbol{\gamma}$ denote the Lagrangian multipliers, and $\mu$, $\tau$ represent the penalty parameters, respectively.
Eqn.~(\ref{equ:MBACFLag}) can be solved iteratively with the ADMM algorithm, in which all the subproblems, i.e., $\mathbf{f}$, $\mathbf{g}$ and $\mathbf{z}$, have their closed-form solutions.
In the following, we present the solution of each subproblem.

\noindent\textbf{Subproblem $\mathbf{g}$}:
\begin{equation}\label{equ:subproblemf}
   \small
   \begin{split}
      \mathbf{g} \!= \mathop{\arg}\min_{\mathbf{g}} \frac{\lambda}{2} \left \| \mathbf{g} \right \|^2 \!+\! \sum_{l=1}^L \boldsymbol{\zeta}_l ^{\text{T}}(\mathbf{f}_l \!-\! \mathbf{P}^{\text{T}}\mathbf{g}_l) \!+\! \frac{\mu}{2}\!\sum_{l=1}^L \left \| \mathbf{f}_l\! -\! \mathbf{P}^{\text{T}}\mathbf{g}_l \right \|^2.
   \end{split}
\end{equation}
Note that each channel of $\mathbf{g}$ in Eqn.~(\ref{equ:subproblemf}) can be computed independently,
thus the closed-form solution of the $l$-th channel of $\mathbf{g}$ can be expressed as,
\begin{equation}\label{equ:solutionf}
   \small
   \mathbf{g}_l = \left(\lambda\mathbf{I} + \mu\mathbf{P}\mathbf{P}^{\text{T}} \right)^{-1} \left(\mathbf{P}\boldsymbol{\zeta}_l + \mu \mathbf{P}\mathbf{f}_l \right),
 \end{equation}
where $\mathbf{I}$ denotes an identity matrix. 
Note that $\lambda \mathbf{I} + \mu \mathbf{P}\mathbf{P}^{\text{T}}$ is a diagonal matrix and its inverse matrix can be efficiently computed via element-wise operation.

\noindent\textbf{Subproblem $\mathbf{f}$}:
\begin{equation}\label{equ:subproblemg}
   \small
   \begin{split}
      \mathbf{f}\!=\! \mathop{\arg}\min_{\mathbf{f}}&\!\sum_{l=1}^L \boldsymbol{\zeta}_l ^{\text{T}}(\mathbf{f}_l \!-\! \mathbf{P}^{\text{T}}\mathbf{g}_l) \!+\! \frac{\mu}{2}\sum_{l=1}^L \left \| \mathbf{f}_l\! -\! \mathbf{P}^{\text{T}}\mathbf{g}_l \right \|^2 \\
       &+\!\boldsymbol{\gamma}^{\text{T}}\!\!\left ( \sum_{l=1}^L \mathbf{x}_{l} \! \star \!\mathbf{f}_l \!-\! \mathbf{y}\!\! -\! \mathbf{z}\! \right )\!\! + \!\frac{\tau}{2}\!\left \| \sum_{l=1}^L \!\mathbf{x}_{l} \! \star \!\mathbf{f}_l\! -\! \mathbf{y}\!\! -\! \mathbf{z} \right \|^2\!.
   \end{split}
\end{equation}
Using Parseval's theorem, Eqn.~(\ref{equ:subproblemg}) can be equivalently expressed in the Fourier domain,
\begin{equation}\label{equ:subproblemgf}
   \small
   \begin{split}
      \hat{\mathbf{f}}= \mathop{\arg}\min_{\hat{\mathbf{f}}}&\sum_{l=1}^L \hat{\boldsymbol{\zeta}}_l ^{\text{T}}\left(\hat{\mathbf{f}}_l \!-\! \hat{\mathbf{q}}_l\right) + \frac{\mu}{2}\sum_{l=1}^L \left \| \hat{\mathbf{f}}_l\! -\! \hat{\mathbf{q}}_l \right \|^2 \\
       &+\!\hat{\boldsymbol{\gamma}}^{\text{T}}\!\!\left ( \sum_{l=1}^L \hat{\mathbf{x}}_{l} \! \odot \!\hat{\mathbf{f}}_l \!-\! \hat{\mathbf{y}}\!\! -\! \hat{\mathbf{z}}\! \right )\!\! + \!\frac{\tau}{2}\left \| \sum_{l=1}^L \!\hat{\mathbf{x}}_{l} \! \odot \!\hat{\mathbf{f}}_l\! -\! \hat{\mathbf{y}}\!\! -\! \hat{\mathbf{z}} \right \|^2\!.
   \end{split}
\end{equation}
Here $\hat{\mathbf{x}} = \sqrt{T}\mathbf{F}\mathbf{x}$ represents the FFT of sample $\mathbf{x}$ where $\mathbf{F}$ is the orthonormal Discrete
Fourier Transform (DFT) matrix, and {$\hat{\mathbf{q}}_l$ takes the form of $\hat{\mathbf{q}}_l = \sqrt{T}\mathbf{FP}^{\text{T}}\mathbf{g}_l$}.
Analogous to BACF~\cite{Galoogahi2017Learning}, the solution for $\hat{\mathbf{f}}$ can be divided into $T$ independent subproblems.
Denote by $\mathbf{x}\left(t\right) \in \mathbb{R}^L$ the vector consisting of $t$-th elements of sample $\mathbf{x}$ along all $L$ channels,
then the $t$-th elements $\hat{\mathbf{f}}(t)$ of $\hat{\mathbf{f}}$ can be computed by,
\begin{equation}\label{equ:solutiong}
   \small
   \begin{split}
      \hat{\mathbf{f}}(t) = &\left(\tau\hat{\mathbf{x}}\left(t\right) \hat{\mathbf{x}}\left(t\right)^{\text{T}} + \mu\mathbf{I}\right)^{-1} \\
      & \left(\!\tau\hat{\mathbf{x}}\left(t\right)\hat{\mathbf{y}}\left(t\right) \!+\! \tau\hat{\mathbf{x}}\left(t\right)\hat{\mathbf{z}}\left(t\right) \!-\!\hat{\mathbf{x}}\left(t\right)\boldsymbol{\gamma}{\left(t\right)}\! - \!\hat{\boldsymbol{\zeta}}\left(t\right) \!+\! \mu\hat{\mathbf{q}}\left(t\right)\!\right).
   \end{split}
 \end{equation}
Note that $\hat{\mathbf{x}}\left(t\right) \hat{\mathbf{x}}\left(t\right)^{\text{T}}$ is rank-1 matrix, thus Eqn.~(\ref{equ:solutiong}) can be efficiently solved with Sherman-Morrison formula~\cite{petersen2008matrix},
\begin{equation}\label{equ:solutiong2}
   \small
   \begin{split}
      \hat{\mathbf{f}}\left(t\right) &= \frac{1}{\mu}\left(\!\tau\hat{\mathbf{x}}\left(t\right)\hat{\mathbf{y}}\left(t\right) \!+\! \tau\hat{\mathbf{x}}\left(t\right)\hat{\mathbf{z}}\left(t\right) \!-\!\hat{\mathbf{x}}\left(t\right)\hat{\boldsymbol{\gamma}}{\left(t\right)}\! - \!\hat{\boldsymbol{\zeta}}\left(t\right) \!+\! \mu\hat{\mathbf{q}}\left(t\right)\!\right)\\
       &-\!\frac{\hat{\mathbf{x}}\left(t\right)}{\mu b}\!\left(\!\tau\hat{\mathbf{y}}\!\left(t\right)\!\hat{s}_{\mathbf{x}}\!\left(t\right) \!+\! \tau\hat{\mathbf{z}}\!\left(t\right)\!\hat{s}_{\mathbf{x}}\!\left(t\right) \!-\!\hat{\boldsymbol{\gamma}}{\left(t\right)}\hat{s}_{\mathbf{x}}\!\left(t\right)\! - \!\hat{s}_{\boldsymbol{\zeta}}\!\left(t\right) \!+\! \mu\hat{s}_{\mathbf{q}}\!\left(t\right)\!\right).
   \end{split}
 \end{equation}
where $\hat{s}_{\mathbf{x}}\left(t\right)\!=\! \hat{\mathbf{x}}\!\left(t\right)^{\text{T}}\!\hat{\mathbf{x}}\!\left(t\right)$, $\hat{s}_{\boldsymbol{\zeta}}\left(t\right)\!=\!\hat{\mathbf{x}}\!\left(t\right)^{\text{T}}\!\hat{\mathbf{\boldsymbol{\zeta}}}\!\left(t\right)$, $\hat{s}_{\mathbf{q}}\!\left(t\right)\!=\!\hat{\mathbf{x}}\!\left(t\right)^{\text{T}}\!\hat{\mathbf{q}}\!\left(t\right)$ and $b\!=\! \frac{\mu}{\tau} + \hat{s}_{\mathbf{x}}\left(t\right)$.
And the solution for $\mathbf{f}$ is further obtained with the inverse DFT operation.

\noindent\textbf{Subproblem $\mathbf{z}$}:
\begin{equation}\label{equ:subproblemz}
   \small
   \begin{split}
      \mathbf{z} = \mathop{\arg}\min_{\mathbf{z}} &\frac{1}{2}\left \| \mathbf{M} \odot \mathbf{z} \right \|^2 +\!\boldsymbol{\gamma}^{\text{T}}\!\!\left ( \sum_{l=1}^L \mathbf{x}_{l} \! \star \!\mathbf{f}_l \!-\! \mathbf{y}\!\! -\! \mathbf{z}\! \right )\!\! \\
      &+ \!\frac{\tau}{2} \left \| \sum_{l=1}^L \!\mathbf{x}_{l} \! \star \!\mathbf{f}_l\! -\! \mathbf{y}\!\! -\! \mathbf{z} \right \|^2\!.
   \end{split}
\end{equation}

Analogous to Eqn.~(\ref{equ:subproblemf}), each element in $\mathbf{z}$ can also be computed independently, and its solution is given as,
\begin{equation}\label{equ:subproblemz1}
   \small
   \mathbf{z} = \left( \text{Diag} ({\mathbf{M} \odot \mathbf{M} + \tau\mathbf{1}} ) \right)^{-1} \left({\tau(\mathop{\sum}\limits_{l=1}^L \!\mathbf{x}_{l} \star \mathbf{f}_l\! -\! \mathbf{y})\!\! + \!\boldsymbol{\gamma}}\right),
 \end{equation}
where $\mathbf{1}$ defines a vector where each element equals to $1$, and $\text{Diag}(\cdot)$ constructs a diagonal matrix from a vector.

\noindent\textbf{Lagrangian Update}: The Lagrangian multipliers $\boldsymbol{\zeta}$, $\boldsymbol{\gamma}$ are updated as,
\begin{equation}\label{equ:Lagrangian}
   \begin{split}
      &\boldsymbol{\zeta}^{(t+1)} = \boldsymbol{\zeta}^{(t)} + \mu\left( \mathbf{f}^{(t+1)} - \mathbf{P}^{\text{T}}\mathbf{g}^{(t+1)} \right), \\
      &\boldsymbol{\gamma}^{(t+1)} = \boldsymbol{\gamma}^{(t)} + \tau\left( \sum_{l=1}^L \mathbf{x}_l \star \mathbf{f}_l^{(t+1)} - \mathbf{y} - \mathbf{z}^{(t+1)} \right).
   \end{split}
 \end{equation}
where $\mathbf{f}^{(t+1)}$, $\mathbf{g}^{(t+1)}$ and $\mathbf{z}^{(t+1)}$ are the current solutions to the above subproblems at iteration $t + 1$ within the iterative ADMM algorithm.

Finally, we also note that the above solutions can be easily extended to remove cosine window from other CF trackers (e.g., STRCF) with a single base image.

   \subsection{Solution for CF trackers with multiple base images}

   Another category of CF trackers with spatial regularization are defined on multiple base images, which inevitably breaks the circulant structure and generally requires iterative optimization to solve some of the resulting subproblems.
   %
   %
   Several representative trackers in this category include SRDCF~\cite{danelljan2015learning}, CCOT~\cite{Danelljan2016CCOT}, ECO~\cite{Danelljan2016ECO} and UPDT~\cite{bhat2018unveiling}.
   In this subsection, we use ECO as an example to suggest an iterative optimization method for removing cosine window.
   Without loss of generality, our solution can be easily extended to remove cosine window from other CF trackers based on multiple base images (e.g., UPDT~\cite{bhat2018unveiling}).


   In general, the learning algorithm in ECO consists of two stages.
   (i) In the first frame, a sample projection matrix is learned with the CF to reduce the number of feature channels in training samples.
   (ii) In the subsequent frames the projection matrix is fixed and the CFs are further updated with the reduced features.
   To keep consistent with the ECO tracker~\cite{Danelljan2016ECO}, we also define the formulation for data on a one-dimension domain.
   Denote by a collection of $K$ sample pairs $\{(\mathbf{x}_k,\mathbf{y}_k)\}_{k=1}^K$, and the feature map size for the $l$-th channel $\mathbf{x}_{k,l} $ is $N_l$.
   The feature map $\mathbf{x}_{k,l}$ in ECO tracker is first transformed into the continuous spatial domain $t \in [0,T)$ with an interpolation operator $J_l$,
   \begin{equation}\label{equ:ECOinterp}
      J_l\{\mathbf{x}_{k,l}\}(t) = \sum_{n=0}^{N_l-1}\mathbf{x}_{k,l}[n]b_l(t-\frac{T}{N_l}n),
    \end{equation}
   where $b_l$ is an interpolation kernel with the period $T > 0$.
   Suppose the reduced correaltion filter $\mathbf{f} = [\mathbf{f}_1,...,\mathbf{f}_D]$ consists of $D$ feature maps with $D < L$, and the sample projection matrix $\mathbf{Q} \in \mathbb{R}^{L \times D}$ is represented with $\mathbf{Q} = \left(q_{l,d}\right)$.
   Then the filter $\mathbf{f}$ and sample projection matrix $\mathbf{Q}$ can be computed by minimizing the following objective function,
   \begin{equation}\label{equ:ECO}
      \small
      \begin{split}
      \mathcal{E}(\mathbf{f},\mathbf{Q}) &= \frac{1}{2}\!\sum_{k=1}^K \!\alpha_k\! \left \| \sum_{d=1}^D \sum_{l=1}^L q_{l,d}\mathbf{f}_d \star \!\left(J_l\{\mathbf{x}_{k,l}\} \odot \mathbf{c}\right)\! -\! \mathbf{y}_k \right \|^2\! \\
      &+ \frac{1}{2}\!\sum_{d=1}^D\left \| \mathbf{w} \odot \mathbf{f}_d \right \|^2\! \!\!+ \frac{\lambda}{2}\!\left \| \mathbf{Q} \right \|^2,
      \end{split}
    \end{equation}
   where $\mathbf{w}$ denotes the spatial regularization matrix.

   When removing cosine window and incorporating the mask function $\mathbf{M}$, the ECO model can be modified as,
   \begin{equation}\label{equ:ECOM}
      \small
      \begin{split}
      \mathcal{E}(\mathbf{f},\mathbf{Q}) \!&= \!\frac{1}{2}\!\sum_{k=1}^K \!\alpha_k\! \left \| \mathbf{M} \odot (\sum_{d=1}^D \sum_{l=1}^L q_{l,d}\mathbf{f}_d \star \!J_l\{\mathbf{x}_{k,l}\} \! -\! \mathbf{y}_k) \right \|^2\! \\
      &+ \frac{1}{2}\!\sum_{d=1}^D\left \| \mathbf{w} \odot \mathbf{f}_d \right \|^2\! \!\!+ \frac{\lambda}{2}\!\left \| \mathbf{Q} \right \|^2.
      \end{split}
    \end{equation}

   To solve Eqn.~(\ref{equ:ECOM}), we introduce a series of auxiliary variables [$\mathbf{z}_1$, ..., $\mathbf{z}_K$] with $\mathbf{z}_k \!=\!\! \sqrt{\alpha_k}(\sum\limits_{d=1}^D \sum\limits_{l=1}^L\! q_{l,d}\mathbf{f}_d \star \!J_l\{\mathbf{x}_{k,l}\} \! - \mathbf{y}_k)$, then it
   can be relaxed as,
   \begin{equation}\label{equ:ECOM2}
      \small
      \begin{split}
      \mathcal{E}(\mathbf{f},\mathbf{Q},\mathbf{z}_k) \!&= \!\frac{1}{2}\!\sum_{k=1}^K  \left \| \mathbf{M} \odot \mathbf{z}_k \right \|^2+ \frac{1}{2}\!\sum_{d=1}^D\left \| \mathbf{w} \odot \mathbf{f}_d \right \|^2\! \!\!+ \frac{\lambda}{2}\!\left \| \mathbf{Q} \right \|^2\\
      &+ \!\frac{\tau}{2}\!\sum_{k=1}^K\alpha_k\left \| \sum_{d=1}^D \sum_{l=1}^L q_{l,d}\mathbf{f}_d \star \!J_l\{\mathbf{x}_{k,l}\} \! -\! \mathbf{y}_k\! - \frac{\mathbf{z}_k}{\sqrt{\alpha_k}} \right\|^2,
      \end{split}
    \end{equation}
   where $\tau$ is a penalty parameter which is updated along with the iterations.

   We suggest an iterative optimization algorithm for solving the problem in Eqn.~(\ref{equ:ECOM2}).
   In particular, we minimize the objective in each iteration by alternating between updating the auxiliary variables $\mathbf{z}_k$ and the model parameters $\{\mathbf{f}$, $\mathbf{Q}\}$, which is further explained as follows.
   %

   \noindent\textbf{Updating $\{\mathbf{f},\mathbf{Q}\}$:} Given the auxiliary variables [$\mathbf{z}_1$, ..., $\mathbf{z}_K$], we can observe that the subproblem shares similar formulation with Eqn.~(\ref{equ:ECO}),
   thus it can be minimized with the optimization method used in the ECO tracker.

   \noindent\textbf{Updating $\mathbf{z}$:} Analogous to Eqn.~(\ref{equ:subproblemz}), the closed-form solution for $\mathbf{z}_k$ can be computed by,
      \begin{equation}\label{equ:ECOMfirstsubproblemz}
      \small
      \mathbf{z}_k \!=\! \left( \text{Diag} ({\mathbf{M} \!\odot\! \mathbf{M} \!+\! \tau\mathbf{1}} ) \right)^{-1}
      {\tau\sqrt{\alpha_k}(\sum_{d=1}^D \sum_{l=1}^L q_{l,d}\mathbf{f}_d \star \!J_l\{\mathbf{x}_{k,l}\} \! -\! \mathbf{y}_k)}.
    \end{equation}

   \section{Experiments}\label{sec:exp}

   In this section, we evaluate the feasibility and effectiveness of removing cosine window by integrating it into five representative CF trackers with spatial regularization, i.e., BACF, STRCF, ECOhc, ECO and UPDT.
   Then, extensive experiments are conducted to compare our methods with the state-of-the-art methods on three popular tracking benchmarks,
   i.e., OTB-2015~\cite{wu2015object}, Temple-Color~\cite{Liang2015Encoding} and VOT-2018~\cite{kristan2018sixth} datasets.

   \subsection{Baseline CF trackers}

   Our methods are generic and can be integrated to multiple CF trackers with spatial regularization, such as those with single or multiple base images, using handcrafted or deep CNN features.
   In the experiments, we choose three baseline CF trackers using handcrafted features, i.e., BACF~\cite{Galoogahi2017Learning}, ECOhc~\cite{Danelljan2016ECO} and STRCF~\cite{li2018learning}.
   Moreover, we also consider two state-of-the-art baseline CF trackers using CNN features, i.e., ECO~\cite{Danelljan2016ECO} and UPDT~\cite{bhat2018unveiling}.
   It is worth noting that we only incorporate our method with UPDT on the VOT-2018 dataset, because UPDT employs the difficult videos from OTB-2015 for parameter tuning and most of these videos also exist in Temple-Color.
   Besides, another two CF trackers without spatial regularization, i.e., MOSSE and KCF, are also included to illustrate when to remove cosine window from CF trackers.

   \subsection{Implementation details}

   We employ the publicly available codes provided by the authors to reproduce the results of the baseline CF trackers and competing methods.
   As for our modified trackers by removing cosine window, we keep most of the parameters the same with their counterparts, and mainly fintune the parameters added by our methods.
   In particular, we set the penalty parameters $\tau$, $\mu$, and the number of iterations in CF$_{RCG}$ as \{2.5, 2.5, 3\}, respectively.
   The penalty parameters $\tau$, $\mu$ are updated along with iterations by $\tau^{(t+1)} = \min(p\tau^{(t)}, \tau_{max})$ and $\mu^{(t+1)} = \min(p\mu^{(t)}, \mu_{max})$,
   where $\tau_{max}$, $\mu_{max}$ and $p$ are set to ${100, 100, 1.05}$, respectively.
   As for the ECO trackers, the parameters $\tau$, the number of iterations are set to \{2.2, 4\} and \{2.5, 5\} for ECOhc$_{RCG}$ and ECO$_{RCG}$, respectively.
   In addition, we assign the standard deviation parameter $\delta$ in Eqn.~(\ref{equ:MDef2}) to \{1.2, 1.4, 2\} for BACF$_{RCG}$, ECOhc$_{RCG}$ and ECO$_{RCG}$, respectively.
   Note that we employ the same parameter settings for each tracker throughout the experiments on all datasets.
   Our method is implemented on Matlab 2017b with Matconvnet library~\cite{vedaldi2015matconvnet}, and all the experiments are run on a PC equipped with Intel i7 7700 CPU, 32GB RAM and a single NVIDIA GTX 1070 GPU.

   \subsection{Internal Analysis of our methods}

   \subsubsection{Ablation study}\label{sec:ablation}

   In this section, we study the effect of removing cosine window, incorporating binary or Gaussian shaped mask functions into the baseline CF trackers using the VOT-2018 benchmark~\cite{kristan2018sixth}.
   To this end, we implement four variants for each baseline CF tracker, i.e., the baseline CF (termed as \textbf{Baseline}), removing cosine window ($\bf{{RC}}$), removing cosine window and incorporating binary mask function ($\bf{{RCB}}$), and removing cosine window and incorporating Gaussian shaped mask function ($\bf{{RCG}}$).
   In addition, we also include MOSSE and KCF as baseline trackers to show that their performance is degraded by removing cosine window and cannot be remedied by incorporating mask function.
   Following the protocols in~\cite{VOTmeasure2016}, we evaluate the performance of each method using Expected Average Overlap (EAO), accuracy and robustness as performance measures.

   Table~\ref{Tab:Ablation} presents the results of all the variants on the VOT-2018 dataset.
   One can observe that when removing the cosine window from MOSSE and KCF, the tracking performance degrades significantly.
   And they still perform inferior to the \textbf{Baseline} counterparts even with the introduction of binary or Gaussian shaped mask functions.
   Thus, cosine window cannot be removed from the CF trackers without spatial regularization.

   \begin{figure}[!htbp]
      \centering
      \includegraphics[width=0.42\textwidth]{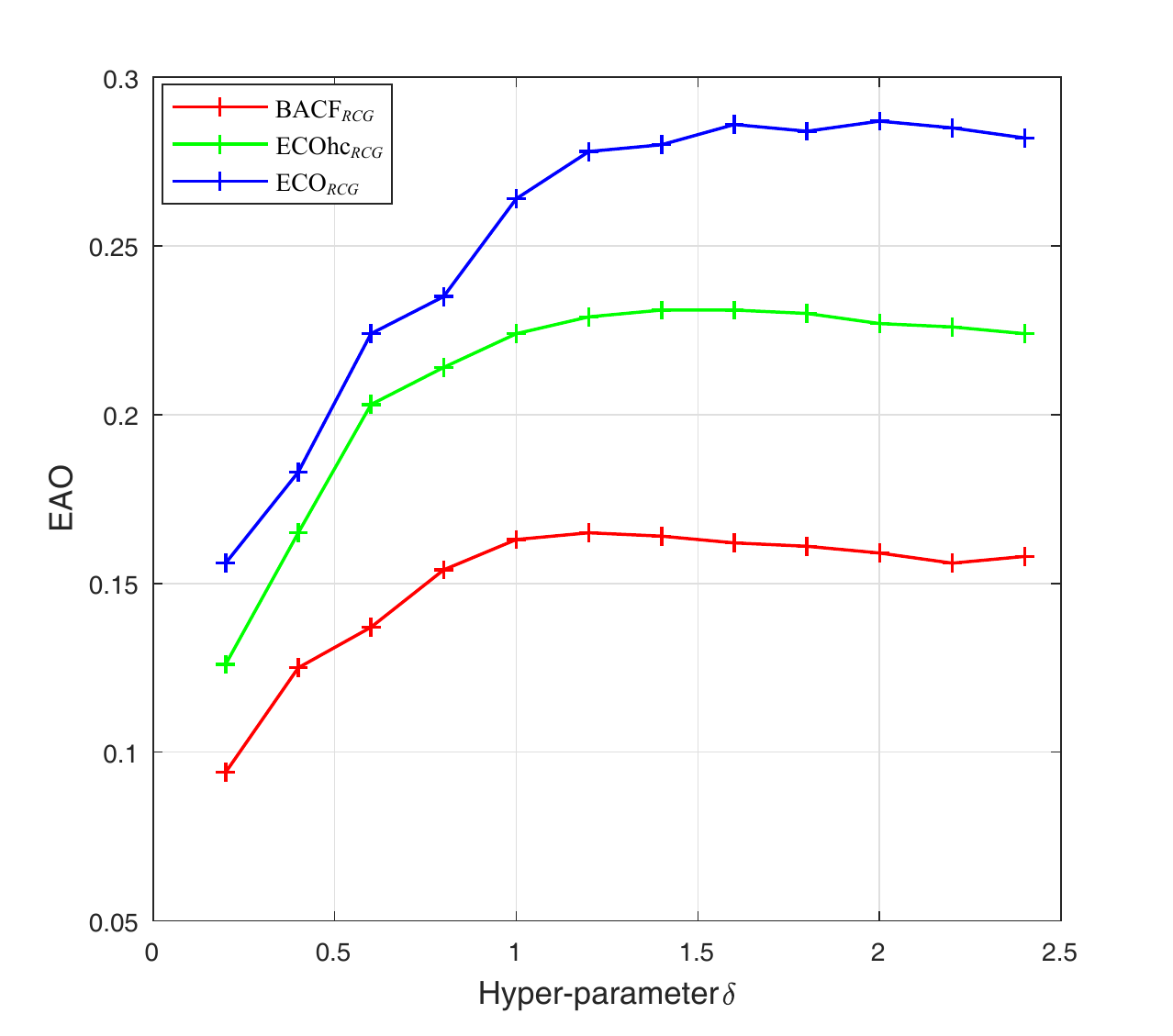}
      \caption{Effect of the hyper-parameter $\delta$ in $\mathbf{M}_G$ for BACF$_{RCG}$, ECOhc$_{RCG}$ and ECO$_{RCG}$ on the VOT-2018 dataset.}
      \label{fig:effectofparams}
   \end{figure}

   As for the CF trackers with spatial regularization, we can make the following observations.
   (i) In comparison to \textbf{Baseline}, the performance of \textbf{RC} variant slightly degrade with a drop of $\sim$1.5 in terms of EAO.
   Such performance degradation can be explained by the fact that a small percentage of negative samples still suffer from boundary discontinuity which may be harmful to tracking performance.
   (ii) By integrating the binary mask function $\mathbf{M}$ into the CF trackers with spatial regularization, the \textbf{RCB} variants consistently outperform the $\textbf{RC}$ and \textbf{Baseline}.
   In terms of EAO, the performance gain of \textbf{RCB} can be about $0.02\sim 0.04$ against $\textbf{RC}$ and about $0.015\sim 0.03$ against \textbf{Baseline}.
   The performance improvement can be ascribed to the reason that \textbf{RCB} is more effective in handling both boundary discontinuity and sample contamination in comparison with \textbf{Baseline} and $\textbf{RC}$.
   %
   %
   (iii) The introduction of Gaussian shaped mask function $\mathbf{M}_G$ can further boost the performance of CF trackers with spatial regularization, indicating that the samples near target center should be emphasized more in the modified CF models.
   (iv) Finally, \textbf{RCB} and \textbf{RCG} significantly improve the robustness against the \textbf{Baseline} trackers with lower failure rate.
   In terms of accuracy, \textbf{RCB} and \textbf{RCG} perform on par with \textbf{Baseline} and \textbf{RC}, indicating that the gain of mask function should be attributed to the improvement on the robustness of appearance modeling.

   To sum up, the results empirically validate our answers to the two problems concerned in this work.
   (i) It is feasible to remove cosine window for CF trackers with spatial regularization.
   (ii) By incorporating with mask function, we can not only safely remove cosine window from CF trackers with spatial regularization, but also bring moderate performance gains over their \textbf{Baseline} counterparts.

   %
   \begin{table*}[!htb]
      \renewcommand\arraystretch{1.6}
      \caption{The EAO, accuracy (Acc.) and robustness (RO.) results by progressively integrating our methods into the baseline CF trackers on the VOT-2018 dataset. Here, \textbf{Baseline}, \textbf{RC}, \textbf{RCB}, and \textbf{RCG} respectively represent
      the baseline CF tracker, that by removing cosine window, that by removing cosine window and incorporating with binary mask function $\mathbf{M}$, and that by removing cosine window and incorporating with Gaussian shaped mask function $\mathbf{M}_G$.
      ($^*$) Note that the results of ECO and UPDT are reproduced from the released codes on the VOT-2018 challenge website, and we report the UPDT results as the average scores of 15 times running following the protocols in~\cite{kristan2018sixth}.}

      \scalebox{0.65}{
      \begin{tabular}{|c||ccc|ccc|ccc|ccc|ccc|ccc|ccc|}
      \hline
      \rowcolor{lightgray}
      \textbf{Methods}  &\multicolumn{3}{>{\columncolor{lightgray}}c|}{\textbf{MOSSE}~\cite{bolme2010visual}} & \multicolumn{3}{>{\columncolor{lightgray}}c|}{\textbf{KCF}~\cite{henriques2015high}}  & \multicolumn{3}{>{\columncolor{lightgray}}c|}{\textbf{BACF}~\cite{Galoogahi2017Learning}}  &  \multicolumn{3}{>{\columncolor{lightgray}}c|}{\textbf{STRCF}~\cite{li2018learning}} &  \multicolumn{3}{>{\columncolor{lightgray}}c|}{\textbf{ECOhc}~\cite{Danelljan2016ECO}} &  \multicolumn{3}{>{\columncolor{lightgray}}c|}{\textbf{ECO}$^*$~\cite{Danelljan2016ECO}} &  \multicolumn{3}{>{\columncolor{lightgray}}c|}{\textbf{UPDT}$^*$~\cite{bhat2018unveiling}} \\ \hline\hline
                        &\textbf{EAO}&\textbf{ACC.}&\textbf{RO.}&\textbf{EAO}&\textbf{ACC.}&\textbf{RO.}&\textbf{EAO}&\textbf{ACC.}&\textbf{RO.}&\textbf{EAO}&\textbf{ACC.}&\textbf{RO.}&\textbf{EAO}&\textbf{ACC.}&\textbf{RO.}&\textbf{EAO}&\textbf{ACC.}&\textbf{RO.}&\textbf{EAO}&\textbf{ACC.}&\textbf{RO.} \\ \hline
      \textbf{Baseline} & 0.067 & 0.387 & 1.862    & 0.106 &0.327 & 1.182   & 0.137 &0.432& 0.757   & 0.174 &0.47& 0.632          & 0.212 &0.524& 0.492          & 0.262 &0.458& 0.323        & 0.352 &0.523& 0.207         \\
      \rowcolor{tinygray}
      \textbf{RC} & 0.033 &0.403& 2.438   & 0.069 &0.374& 1.823    & 0.124 &0.466& 0.892              & 0.166 &0.486& 0.683          & 0.194 &0.532& 0.521          & 0.251 &0.476& 0.334        & 0.343 &0.529& 0.221         \\
      \textbf{RCB} & 0.038 &0.408& 2.392  & 0.075 &0.377& 1.788   & 0.158 &0.462& 0.723         & 0.187 &0.478& 0.604          & 0.225 &0.528& 0.476          & 0.279 &0.464& 0.272        & 0.384 &0.524& 0.174         \\
      \rowcolor{tinygray}
      \textbf{RCG} & 0.042 &0.396& 2.375   & 0.081 &0.372& 1.774   & 0.165 &0.458& 0.692        & 0.192 &0.474& 0.595         & 0.231 &0.524& 0.464          & 0.287 &0.462& 0.258       & 0.391 &0.528& 0.168         \\ \hline
      \end{tabular}}
      \vspace{0.01in}
      \label{Tab:Ablation}
   \end{table*}

   \subsubsection{Effect of hyper-parameter $\delta$ in $\mathbf{M}_G$} \label{sec:effectofparams}

   The hyper-parameter $\delta$ in Gaussian shaped mask function $\mathbf{M}_G$ controls the decay speed of training samples from target center to boundaries.
   In particular, higher $\delta$ indicates the slower decay speed, and more negative samples near boundary will be considered during training.
   When $\delta \to +\infty$, the Gaussian shaped mask function $\mathbf{M}_G$ degrades to the binary mask function $\mathbf{M}$.
   Using BACF, ECOhc and ECO, we analyze the effect of the hyper-parameter $\delta$ on tracking performance.
   Concretely, Fig.~\ref{fig:effectofparams} show the EAO plots of the three trackers with different $\delta$ values on the VOT-2018 dataset.
   It can seen that the choice of $\delta$ has a significant effect on EAO score for all the three trackers.
   For BACF, ECOhc and ECO, the \textbf{RCG} methods achieve the best performance when $\delta = \{1.2, 1.4, 2\}$, respectively.

   \subsection{VOT-2018 benchmark}

   To further assess the proposed methods, we compare our best trackers (i.e., UPDT$_{RCG}$ and ECO$_{RCG}$) with the state-of-the-art trackers on the VOT-2018 dataset.
   VOT-2018 consists of 60 challenging videos collected from real-life datasets.
   In the benchmark, a tracker will be re-initialized with the ground-truth bounding boxes whenever it significantly drifts from the target.
   And the performance is evaluated with three measures: accuracy, robustness and EAO.
   The accuracy computes the average overlap between estimated bounding boxes and ground-truth annotations.
   The robustness score counts the times of tracking failures.
   And EAO measure is a principled combination of accuracy and robustness scores.

   Table~\ref{Tab:VOT} lists the results of our UPDT$_{RCG}$ and ECO$_{RCG}$, ECO, and the top ten best performing trackers on the VOT-2018 challenge.
   For a fair comparison, we reproduce the results of ECO and UPDT with their publicly available codes on the VOT-2018 challenge website, and the UPDT result is reported as the average score of 15 times running.
   We also note that the reported EAO score of UPDT on the VOT-2018 challenge is 0.378, while our reproducing result is 0.352 based on the released code on the VOT-2018 challenge.
   From Table~\ref{Tab:VOT}, we can observe that UPDT$_{RCG}$ slightly outperforms VOT-2018 challenge winner LADCF and ranks the first among all the competing trackers.
   UPDT$_{RCG}$ is also superior to its counterpart UPDT by an EAO gain of 3.9\%, indicating the feasibility and benefit of removing cosine window.
   Not surprisingly, ECO$_{RCG}$ also shows its superiority, i.e., an improvement of 2.5\% by EAO score, over the ECO counterpart.

   \begin{table*}[!htb]
      \renewcommand\arraystretch{1.6}
      \centering
      \caption{Comparison with the state-of-the-art trackers in terms of EAO, accuracy, and robustness on the VOT-2018 dataset.
      The {\color{red}{first}}, {\color{green}{second}} and {\color{blue}{third}} best results are highlighted in color. ($^*$) Note that the results of ECO and UPDT are reproduced from the released codes on VOT-2018 challenge website,
      and we report the UPDT result as the average score of 15 times running following the protocols in~\cite{kristan2018sixth}.}
      \scalebox{0.737}{
      \begin{tabular}{|c||c|c|c|c|c|c|c|c|c|c|c|c|c|}
      \hline
      \rowcolor{lightgray}
      \textbf{Methods}         & \tabincell{c}{\textbf{ECO}$^*$\\~\cite{Danelljan2016ECO}}  & \tabincell{c}{\textbf{DLSTpp}\\~\cite{yang2017deep}} & \tabincell{c}{\textbf{SA\_{Siam}\_R}\\~\cite{he2018twofold}} & \tabincell{c}{\textbf{CPT}\\~\cite{kristan2018sixth}} & \tabincell{c}{\textbf{DeepSTRCF}\\~\cite{li2018learning}} & \tabincell{c}{\textbf{UPDT}$^*$\\~\cite{bhat2018unveiling}} & \tabincell{c}{\textbf{DRT}\\~\cite{sun2018correlation}}  & \tabincell{c}{\textbf{RCO}\\~\cite{kristan2018sixth}}  & \tabincell{c}{\textbf{SiamRPN}\\~\cite{li2018high}}  & \tabincell{c}{\textbf{MFT}\\~\cite{kristan2018sixth}}  & \tabincell{c}{\textbf{LADCF}\\~\cite{xu2018learning}}  & \tabincell{c}{\textbf{ECO}$_{RCG}$\\Ours} & \tabincell{c}{\textbf{UPDT}$_{RCG}$\\Ours}\\ \hline\hline
      \textbf{EAO ($\uparrow$)}        & 0.262                             & 0.325     &   0.337  & 0.339 & 0.345    & 0.352 & 0.356        & 0.376        & 0.383            & \textbf{\color{blue}{0.385}}        & \textbf{\color{green}{0.389}}         & 0.287                & \textbf{\color{red}{0.391}}                 \\
      \rowcolor{tinygray}
      \textbf{Accuracy ($\uparrow$)}   & 0.458                              & \textbf{\color{blue}{0.543}}   &   \textbf{\color{green}{0.566}}  & 0.506   & 0.523   & 0.523 & 0.519                   & 0.507        & \textbf{\color{red}{0.586}}            & 0.505        & 0.503          & 0.462                & 0.528                 \\
      \textbf{Robustness ($\downarrow$)} & 0.323                            & 0.224      &   0.258 &  0.239 & 0.215   & 0.207  & 0.201           & \textbf{\color{blue}{0.155}}            & 0.276             & \textbf{\color{red}{0.14}}         & \textbf{\color{green}{0.159}}            & 0.258                & 0.168                  \\ \hline
      \end{tabular}}
      \vspace{0.01in}
      \label{Tab:VOT}
   \end{table*}

   \subsection{OTB-2015 dataset}

   The OTB-2015 dataset~\cite{wu2015object} consists of 100 full annotated videos with 11 video attributes,
   including illumination variations (IV), scale variation (SV), occlusion (OCC), in-plane rotation (IPR), out-of-plane rotation (OPR),
   motion blur (MB), fast motion (FM), deformation (DEF), background clutter (BC), out of view (OV) and low resolution (LR).
   Following the settings given in~\cite{wu2015object}, we evaluate the trackers based on the One Pass Evaluation (OPE) protocol,
   and adopt the overlap precision (OP) metric for calculating the fraction of frames with bounding box overlaps exceeding 0.5 in a sequence.
   Besides, we also present the overlap success plots with different overlap thresholds for detailed comparison.

   To assess our methods, we compare four of them (i.e., STRCF$_{RCG}$, ECOhc$_{RCG}$, BACF$_{RCG}$ and ECO$_{RCG}$) with 22 state-of-the-art trackers,
   which can be roughly grouped into two categories:
   (i) trackers using handcrafted features (i.e., STRCF~\cite{li2018learning}, ECOhc~\cite{Danelljan2016ECO}, BACF~\cite{Galoogahi2017Learning}, DSST~\cite{danelljan2016discriminative}, SAMF$_{AT}$~\cite{bibi2016target}, Staple~\cite{bertinetto2015staple}, TRACA~\cite{ChoiCVPR2018}, SRDCFDecon~\cite{Danelljan2016Adaptive}, SRDCF~\cite{danelljan2015learning}, SKSCF~\cite{zuo2016learning}),
   and (ii) trackers using deep CNN features (i.e., ECO~\cite{Danelljan2016ECO}, CCOT~\cite{Danelljan2016CCOT}, MDNet~\cite{NamH16}, CNN-SVM~\cite{hong2015online}, FCNT~\cite{wang2015visual}, CF-Net~\cite{valmadre2017end}, DeepSTRCF~\cite{li2018learning}, VITAL~\cite{song2018vital}, DeepSRDCF~\cite{danelljan2015convolutional}, SiameseFC~\cite{bertinetto2016fully},
   HDT~\cite{qi2016hedged} and HCF~\cite{ma2015hierarchical}).
   In particular, STRCF$_{RCG}$, ECOhc$_{RCG}$, and BACF$_{RCG}$ are compared with the trackers using handcrafted features, while ECO$_{RCG}$ is compared with the trackers using deep CNN features.
   For a fair comparison, UPDT and UPDT$_{RCG}$ are not included in the comparison because UPDT adopts the difficult videos from OTB-2015 for parameter tuning.

   \begin{table*}[!htb]
      \renewcommand\arraystretch{1.6}
      \centering
      \caption{The mean OP results (\%) of different trackers using handcrafted and CNN features on the OTB-2015 dataset. Note that the first two rows compare the methods with handcrafted features, while the last two rows correspond to the trackers with CNN features.
      The {\color{red}{first}}, {\color{green}{second}} and {\color{blue}{third}} best results are highlighted in color.}
      \scalebox{0.652}{
      \begin{tabular}{|c||c|c|c|c|c|c|c|c|c|c|c|c|c|c|c|c|}
      \hline
      \rowcolor{lightgray}
      \textbf{Methods}             & \tabincell{c}{\textbf{DSST}\\~\cite{danelljan2016discriminative}} & \tabincell{c}{\textbf{SKSCF}\\~\cite{zuo2016learning}} &  \tabincell{c}{\textbf{SAMF}$_{AT}$\\~\cite{bibi2016target}} & \tabincell{c}{\textbf{Staple}\\~\cite{bertinetto2015staple}} & \tabincell{c}{\textbf{SRDCF}\\~\cite{danelljan2015learning}} & \tabincell{c}{\textbf{TRACA}\\~\cite{ChoiCVPR2018}} & \tabincell{c}{\textbf{SRDCFDecon}\\~\cite{Danelljan2016Adaptive}} & \tabincell{c}{\textbf{BACF}\\~\cite{Galoogahi2017Learning}} & \tabincell{c}{\textbf{ECOhc}\\~\cite{Danelljan2016ECO}} & \tabincell{c}{\textbf{STRCF}\\~\cite{li2018learning}}  & \tabincell{c}{\textbf{BACF}$_{RCG}$\\Ours} & \tabincell{c}{\textbf{ECOhc}$_{RCG}$\\Ours} & \tabincell{c}{\textbf{STRCF}$_{RCG}$\\Ours}\\ \hline\hline
       \textbf{Mean OP ($\uparrow$)}  & 62.2 & 66.5   & 68   & 71     & 72.8     & 74.7      & 76.6                & 76.7               & 77.2           & \textbf{\color{blue}{80}}              & 77.8                 & \textbf{\color{green}{78.5}}                   & \textbf{\color{red}{82.3}}                   \\ \hline \hline
       \rowcolor{lightgray}
       \textbf{Methods}              & \tabincell{c}{\textbf{CNN-SVM}\\~\cite{hong2015online}} & \tabincell{c}{\textbf{HCF}\\~\cite{ma2015hierarchical}} & \tabincell{c}{\textbf{HDT}\\~\cite{qi2016hedged}} & \tabincell{c}{\textbf{FCNT}\\~\cite{wang2015visual}} & \tabincell{c}{\textbf{SiameseFC}\\~\cite{bertinetto2016fully}} & \tabincell{c}{\textbf{CF-Net}\\~\cite{valmadre2017end}} & \tabincell{c}{\textbf{DeepSRDCF}\\~\cite{danelljan2015convolutional}} & \tabincell{c}{\textbf{CCOT}\\~\cite{Danelljan2016CCOT}} & \tabincell{c}{\textbf{ECO}\\~\cite{Danelljan2016ECO}}  & \tabincell{c}{\textbf{DeepSTRCF}\\~\cite{li2018learning}} & \tabincell{c}{\textbf{MDNet}\\~\cite{NamH16}} & \tabincell{c}{\textbf{VITAL}\\~\cite{song2018vital}} & \tabincell{c}{\textbf{ECO}$_{RCG}$\\Ours}\\ \hline\hline
      \textbf{Mean OP ($\uparrow$)} & 65.1 & 65.6           & 65.8     & 67.1      & 71       &73           & 76.8                 & 82.4           & 84.8      & \textbf{\color{blue}{84.9}}         & \textbf{\color{blue}{84.9}}          & \textbf{\color{green}{86.6}}                           & \textbf{\color{red}{86.7}}                   \\ \hline
      \end{tabular}}
      \vspace{0.01in}
      \label{Tab:OTB}
      \end{table*}

   \begin{table*}[!htb]
      \renewcommand\arraystretch{1.6}
      \centering
      \caption{The mean OP results (\%) of different trackers using handcrafted features on each attribute of OTB-2015. The {\color{red}{first}}, {\color{green}{second}} and {\color{blue}{third}} best results are highlighted in color.}
      \scalebox{0.725}{
      \begin{tabular}{|c||c|c|c|c|c|c|c|c|c|c|c|c|c|c|c|c|}
      \hline
      \rowcolor{lightgray}
      \textbf{Methods}            & \tabincell{c}{\textbf{DSST}\\~\cite{danelljan2016discriminative}} & \tabincell{c}{\textbf{SKSCF}\\~\cite{zuo2016learning}} &  \tabincell{c}{\textbf{SAMF}$_{AT}$\\~\cite{bibi2016target}} & \tabincell{c}{\textbf{Staple}\\~\cite{bertinetto2015staple}} & \tabincell{c}{\textbf{SRDCF}\\~\cite{danelljan2015learning}} & \tabincell{c}{\textbf{TRACA}\\~\cite{ChoiCVPR2018}} & \tabincell{c}{\textbf{SRDCFDecon}\\~\cite{Danelljan2016Adaptive}} & \tabincell{c}{\textbf{BACF}\\~\cite{Galoogahi2017Learning}} & \tabincell{c}{\textbf{ECOhc}\\~\cite{Danelljan2016ECO}} & \tabincell{c}{\textbf{STRCF}\\~\cite{li2018learning}}  & \tabincell{c}{\textbf{BACF}$_{RCG}$\\Ours} & \tabincell{c}{\textbf{ECOhc}$_{RCG}$\\Ours} & \tabincell{c}{\textbf{STRCF}$_{RCG}$\\Ours}\\ \hline\hline
         \textbf{MB}  & 55.3 & 63.4   & 70.8   & 65     & 72.9     & 73.9      & \textbf{\color{green}{79.9}}  & 73.5  & 75.3     & \textbf{\color{blue}{79.4}}  & 78.4   & 77.4          &  \textbf{\color{red}{81.5}}                  \\
         \rowcolor{tinygray}
         \textbf{OCC}  & 56.2 & 63.4   & 64.8   & 68     & 67.6     & 71.2      & 73.5    & 71.1     & \textbf{\color{blue}{74.5}}     & \textbf{\color{green}{75.1}}     &73.1 & \textbf{\color{blue}{74.5}}      &  \textbf{\color{red}{80.2}}                 \\
         \textbf{IV}  & 65.8 & 68.6   & 62.9   & 72     & 74.2     & 76.9      & \textbf{\color{blue}{79.3}}   & 78.5    & 76.1   & 78.3     & 78.8  & \textbf{\color{red}{81}}         &  \textbf{\color{green}{80.9}}                \\
         \rowcolor{tinygray}
         \textbf{BC}  & 59.9 & 69   & 63   & 67.7     & 69.2     & 74.1      & 78       & 76         & 76.5 & \textbf{\color{blue}{79.5}}    & \textbf{\color{green}{79.9}}  & 78.6        &  \textbf{\color{red}{82.6}}                 \\
         \textbf{IPR}  & 60.8 & 65.4   & 65.5   & 66.9     & 66.3     & 71.5      & 70    & 71.5          & 68.5  & \textbf{\color{green}{73.9}}  & \textbf{\color{blue}{72.4}}  & 70.8        & \textbf{\color{red}{76}}                   \\
         \rowcolor{tinygray}
         \textbf{OPR}  & 58.3 & 64.3   & 64.8   & 66.5     & 65.9     & 72.5      & 72.4       & 71.8    & 72.1   & \textbf{\color{green}{76.6}}    & 72.2   & \textbf{\color{blue}{74.8}}        & \textbf{\color{red}{80.1}}                   \\
         \textbf{SV}  & 55.8 & 56.3   & 58.8   & 61.5     & 67.1     & 68.6      & \textbf{\color{blue}{74.4}}       & 70.2     & 71.9   & \textbf{\color{green}{76.4}}    &  73 & 73.6        & \textbf{\color{red}{78.1}}                   \\
         \rowcolor{tinygray}
         \textbf{FM}  & 55 & 63.2   & 66.8   & 65.9     & 72.1     & 70.6      & \textbf{\color{blue}{74.6}}        & \textbf{\color{green}{76}}      & 74.5   & \textbf{\color{green}{76}}   & \textbf{\color{blue}{74.6}}   & 72.2         &  \textbf{\color{red}{77}}                  \\
         \textbf{DEF}  & 53.1 & 62.7   & 58.4   & 68.6     & 66.1     & 70      & 68.2    & 71.3    & \textbf{\color{blue}{73.7}} & 73.3   & 68.7  & \textbf{\color{red}{75.4}}      &  \textbf{\color{green}{73.9}}                  \\
         \rowcolor{tinygray}
         \textbf{OV}  & 45.5 & 45.8   & 60.3   & 52.3     & 52.7     & \textbf{\color{blue}{67.8}}       & 61.8   & 67.1   & 63.5    & \textbf{\color{red}{70.9}}   & 63.8 & \textbf{\color{blue}{67.8}}      &  \textbf{\color{green}{70.6}}                  \\
         \textbf{LR}  & 34.7 & 24   & 51.4   & 39.3     & 64.1     & 54.9      & 63.9   & 62.2  & 56 & \textbf{\color{red}{69.6}}   & \textbf{\color{blue}{67}}  & 53.8      & \textbf{\color{green}{68.9}}                    \\ \hline
      \end{tabular}}
      \vspace{0.01in}
      \label{Tab:OTBAttrb}
      \end{table*}

   \subsubsection{Comparison with state-of-the-arts}

\begin{figure}[!htbp]
      \centering
      \subfloat[]{\label{fig:OTBAUCfig(1)}
         \includegraphics[width=0.24\textwidth]{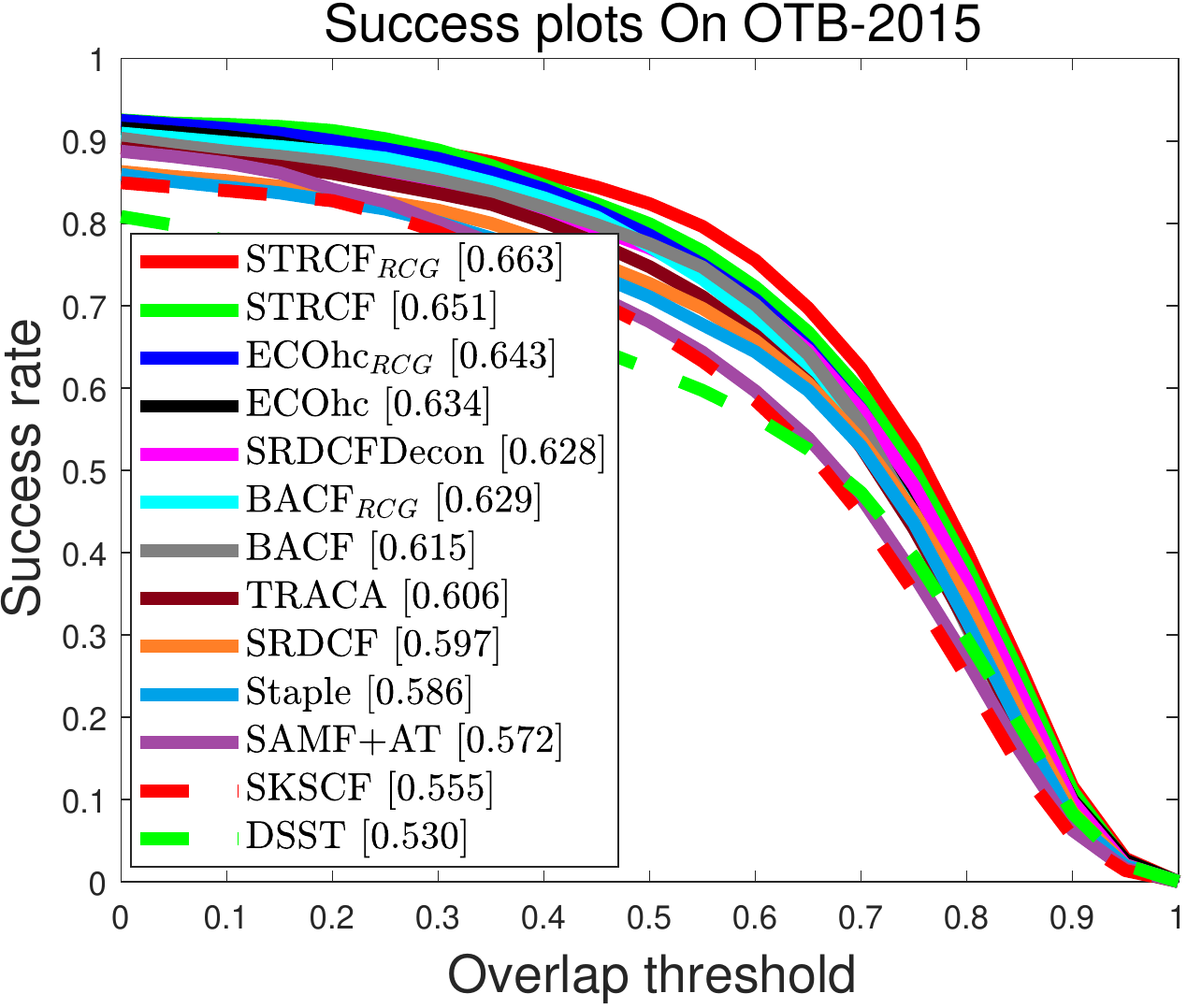}}
      \subfloat[]{\label{fig:OTBAUCfig(2)}
         \includegraphics[width=0.24\textwidth]{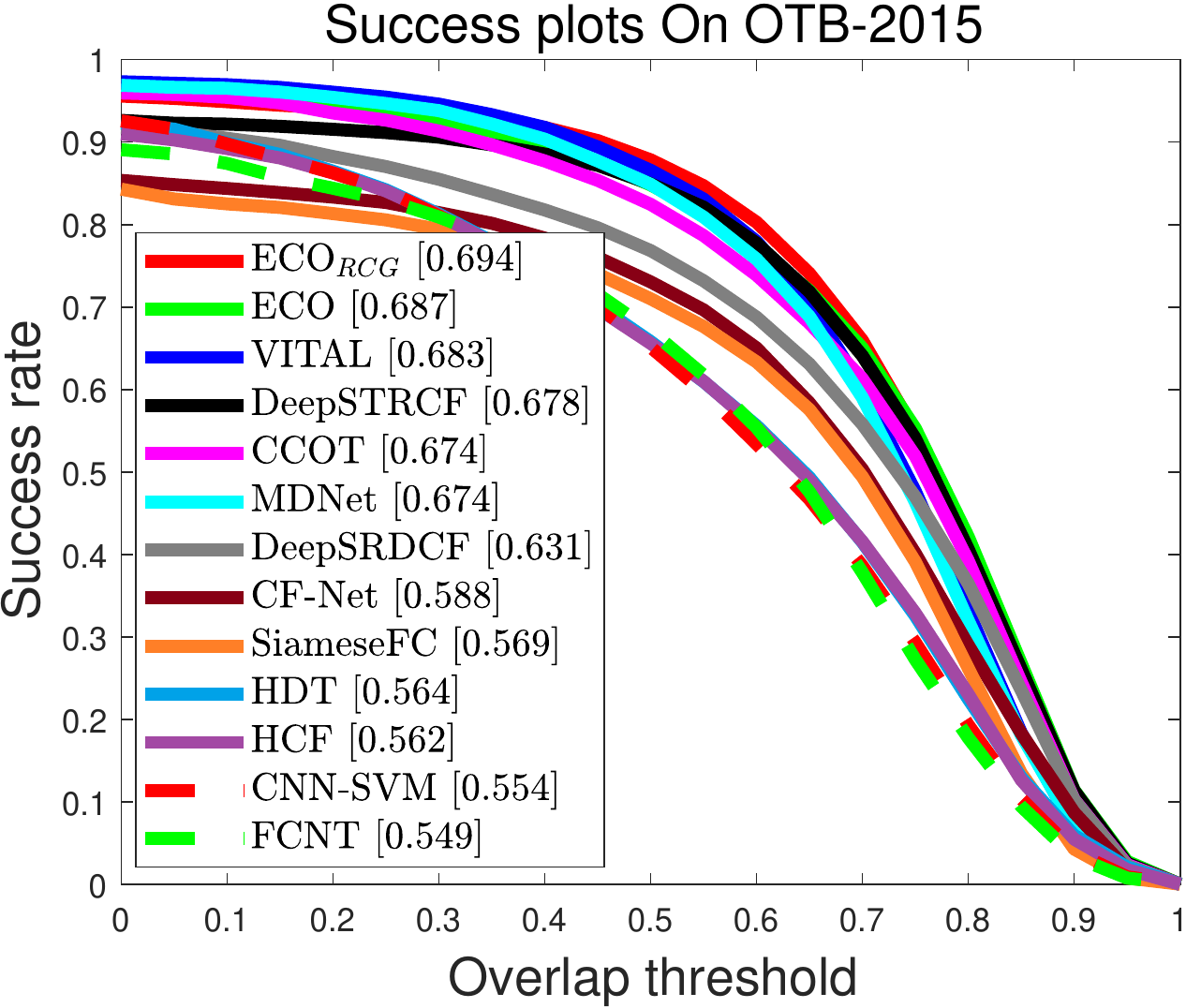}}
      \caption{Comparison of overlap success plots with the state-of-the-art trackers on the OTB-2015 dataset: (a) trackers using handcrafted features, and (b) trackers using deep CNN features.}
      \label{fig:OTB}
      \end{figure}

   We compare the proposed methods with the state-of-the-art trackers on OTB-2015.
   Table~\ref{Tab:OTB} lists the mean OP results of all the competing methods.
   One can see that our methods are consistently superior to their baseline counterparts.
   Using handcrafted features, BACF$_{RCG}$, ECOhc$_{RCG}$ and STRCF$_{RCG}$ outperform their counterparts with mean OP gains of 1.1\%, 1.3\% and 2.3\%, respectively.
   Using deep CNN features, ECO$_{RCG}$ also surpasses its ECO counterpart by 1.9\% in terms of mean OP.
   Moreover, our STRCF$_{RCG}$ achieves the best mean OP among the trackers using handcrafted features, while our ECO$_{RCG}$ performs the best among those using deep CNN features.
   Furthermore, Fig.~\ref{fig:OTB} shows the overlap success curves of the competing methods, which are ranked with the Area-Under-the-Curve (AUC) score.
   Not surprisingly, our methods perform favorably against the competing trackers using handcrafted and deep CNN features.

   \subsubsection{Attribute comparison}

\begin{figure*}[!htbp]
      \centering
      \subfloat[]{\label{fig:MB}
      \includegraphics[width=0.243\textwidth]{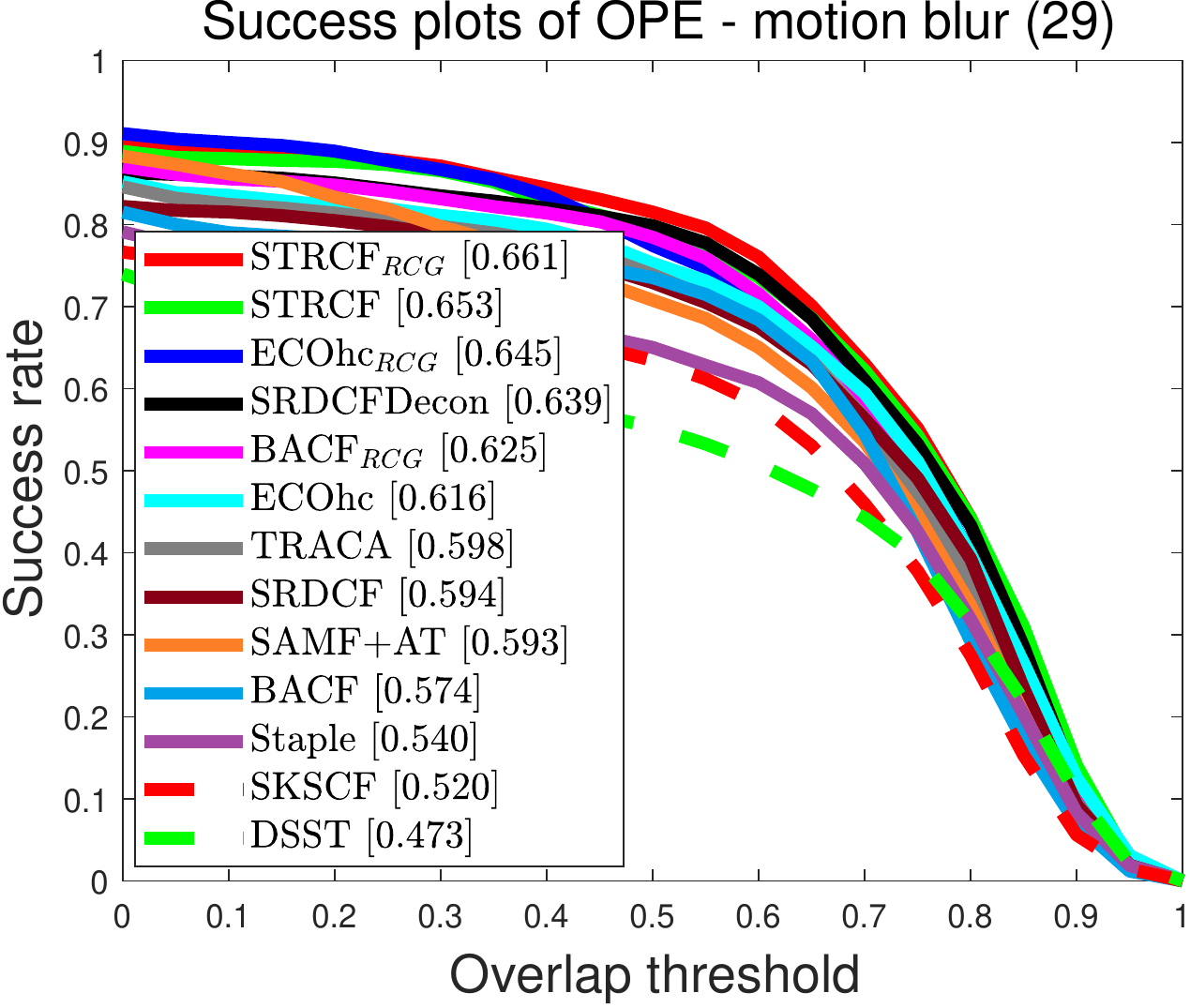}}
      \subfloat[]{\label{fig:OCC}
      \includegraphics[width=0.243\textwidth]{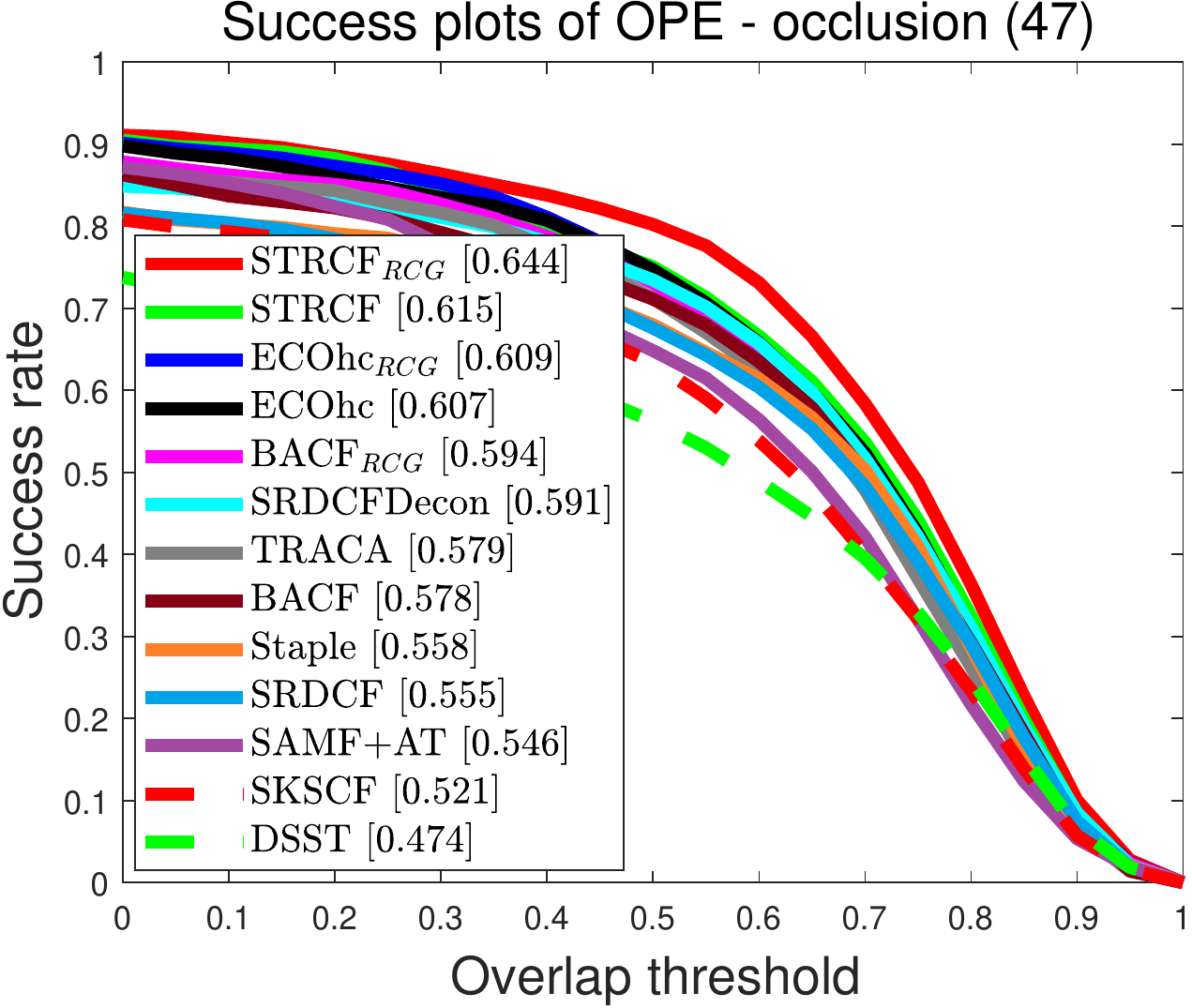}}
      \subfloat[]{\label{fig:IV}
      \includegraphics[width=0.243\textwidth]{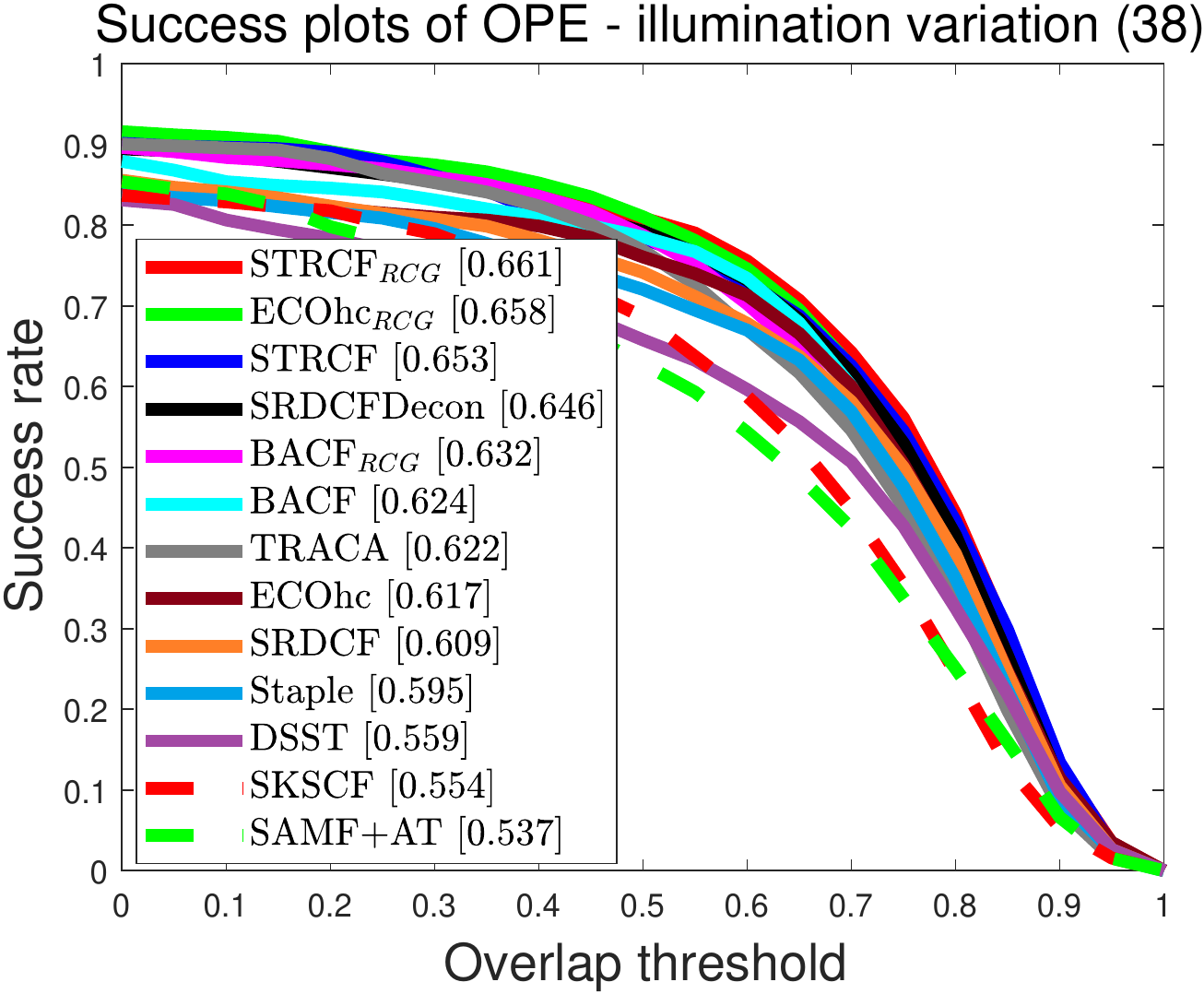}}
      \subfloat[]{\label{fig:BC}
      \includegraphics[width=0.243\textwidth]{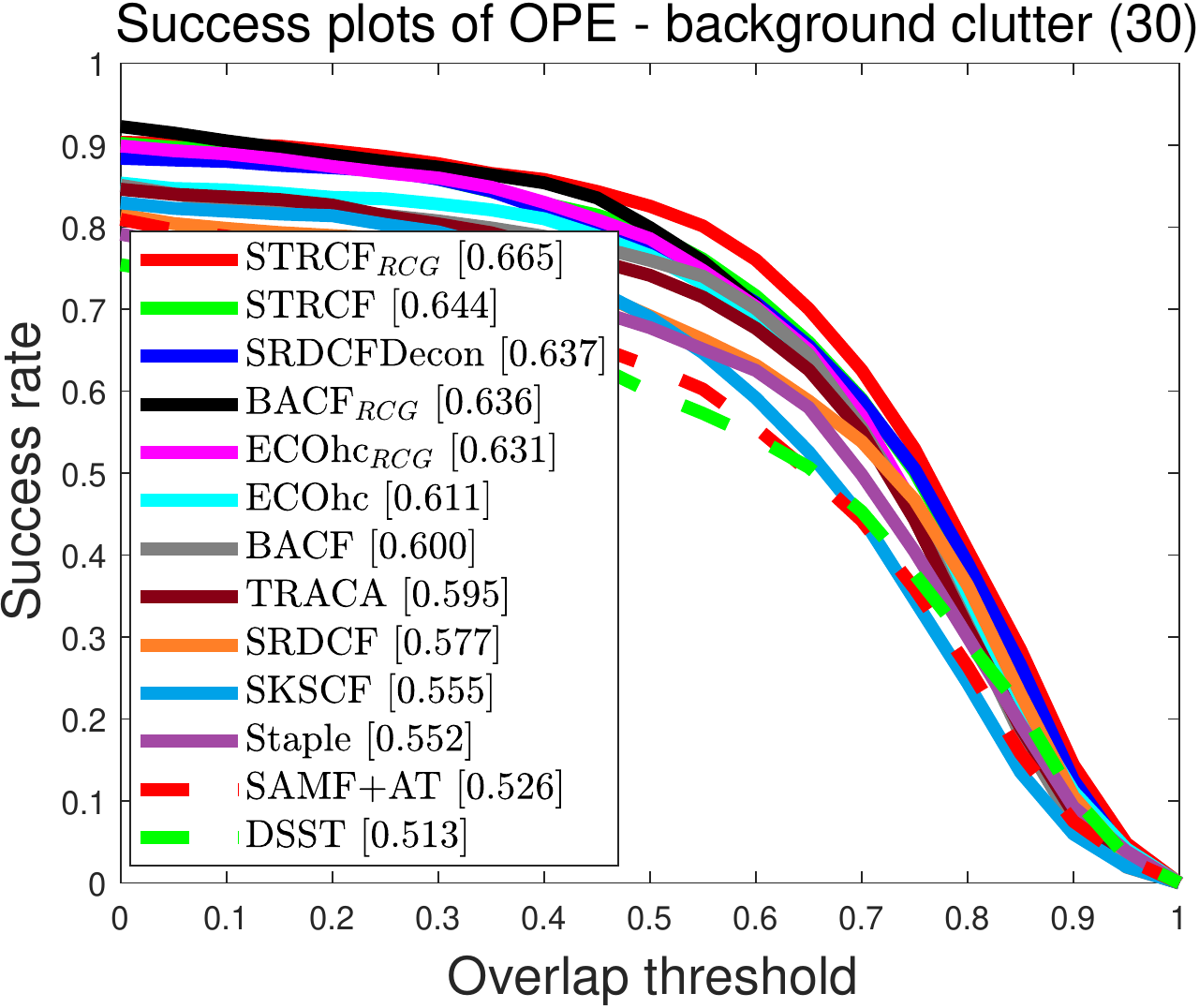}} \\
      \subfloat[]{\label{fig:IPR}
      \includegraphics[width=0.243\textwidth]{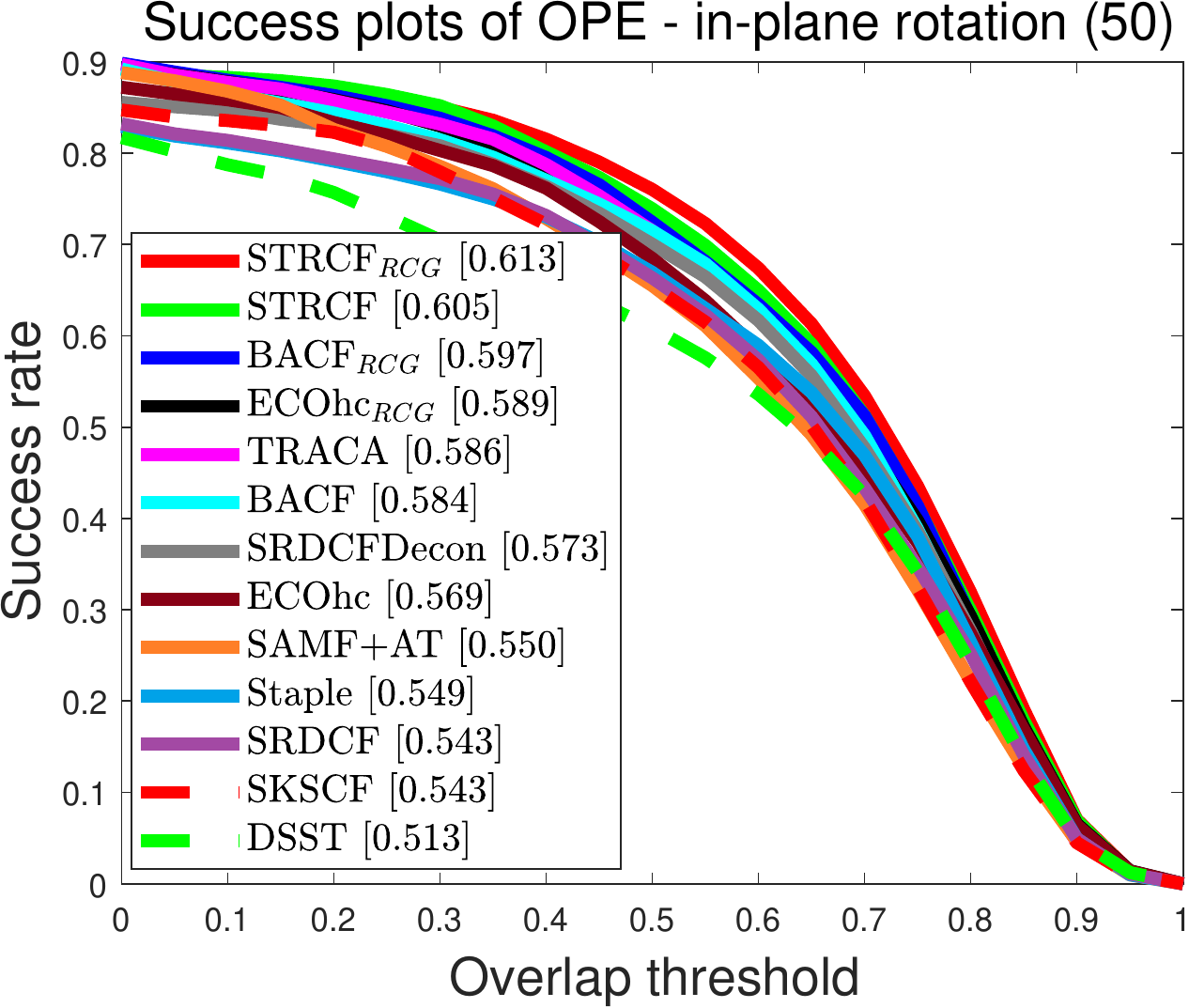}}
      \subfloat[]{\label{fig:OPR}
      \includegraphics[width=0.243\textwidth]{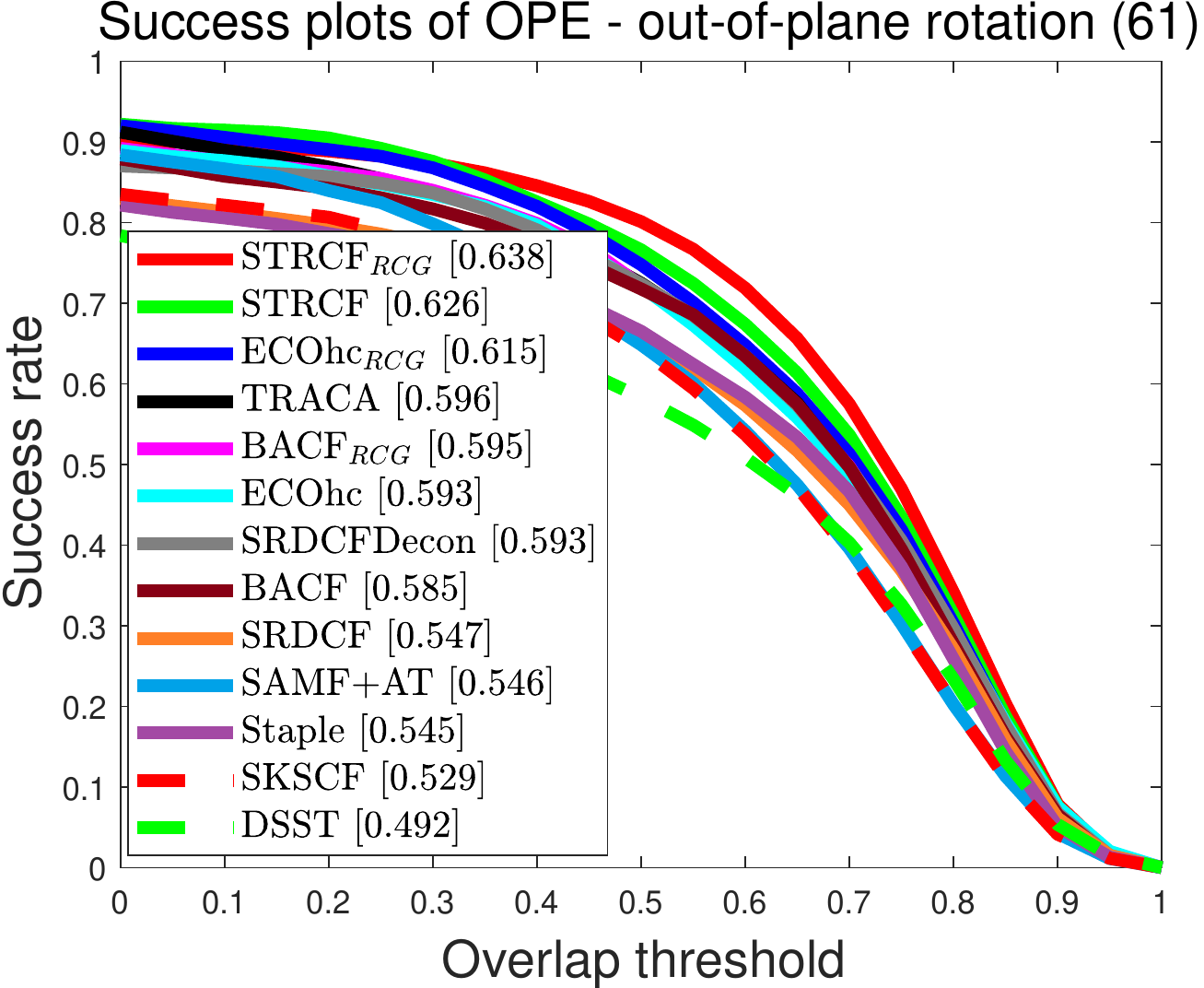}}
      \subfloat[]{\label{fig:SV}
      \includegraphics[width=0.243\textwidth]{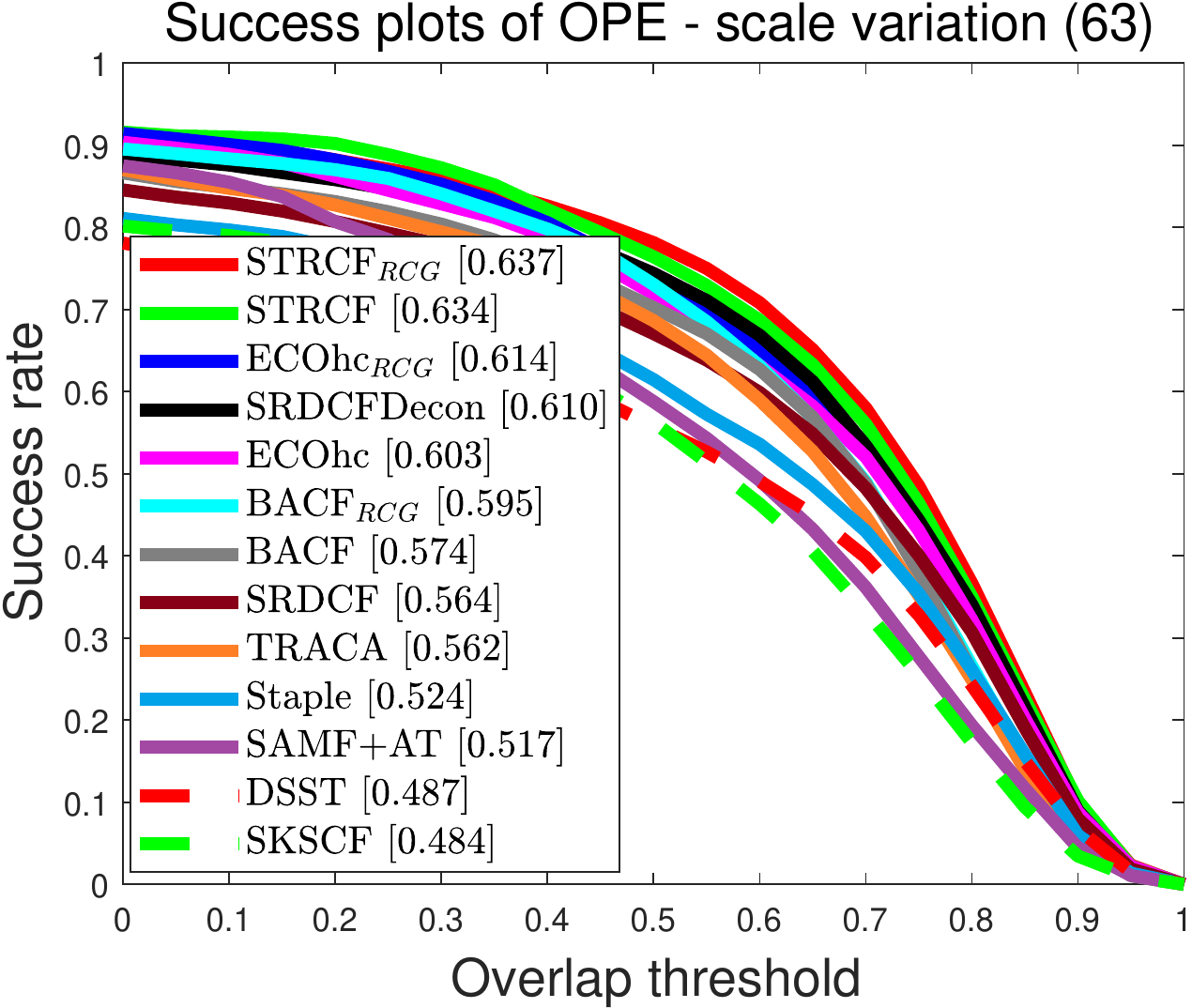}}
      \subfloat[]{\label{fig:FM}
      \includegraphics[width=0.243\textwidth]{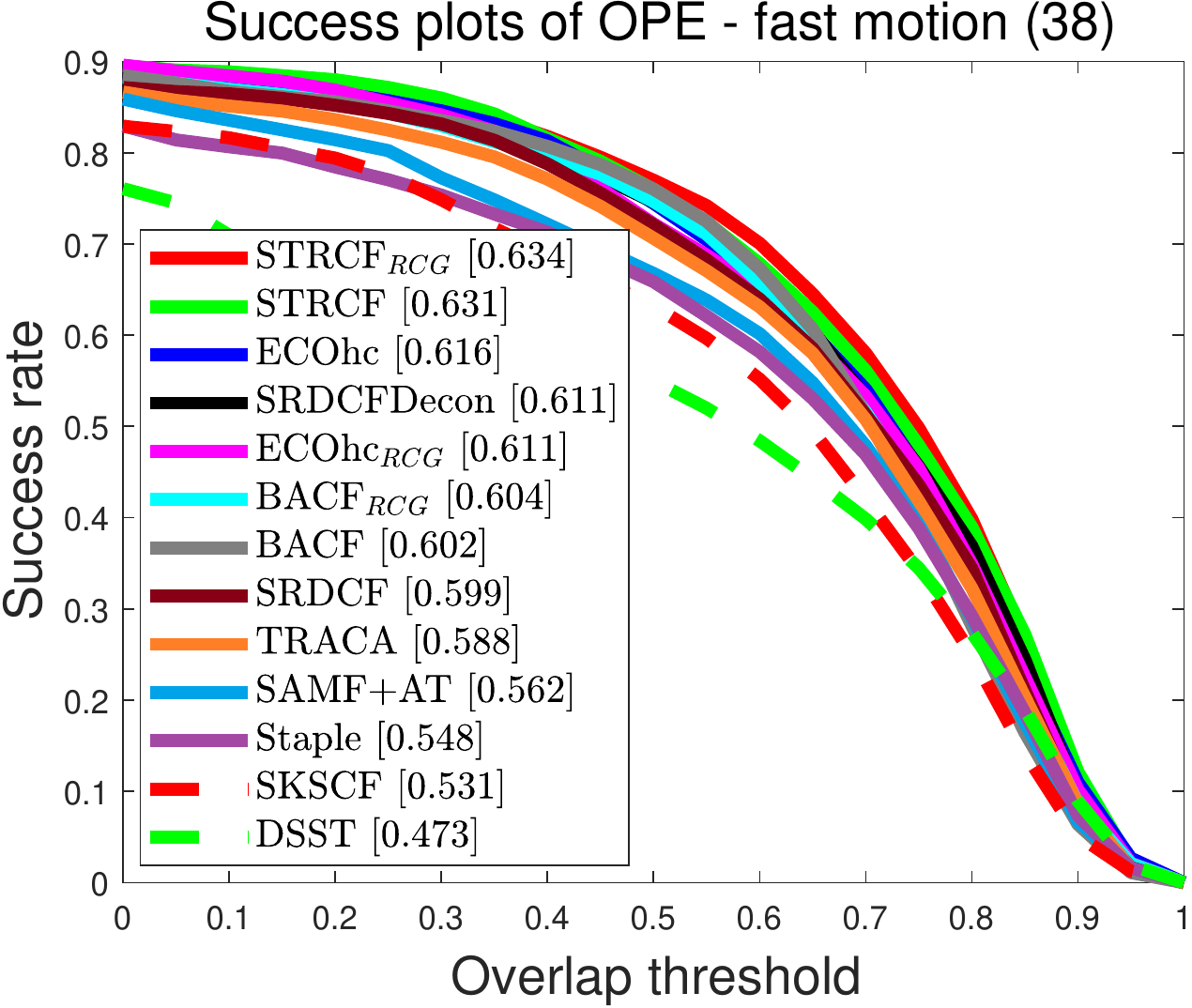}} \\
      \subfloat[]{\label{fig:DEF}
      \includegraphics[width=0.243\textwidth]{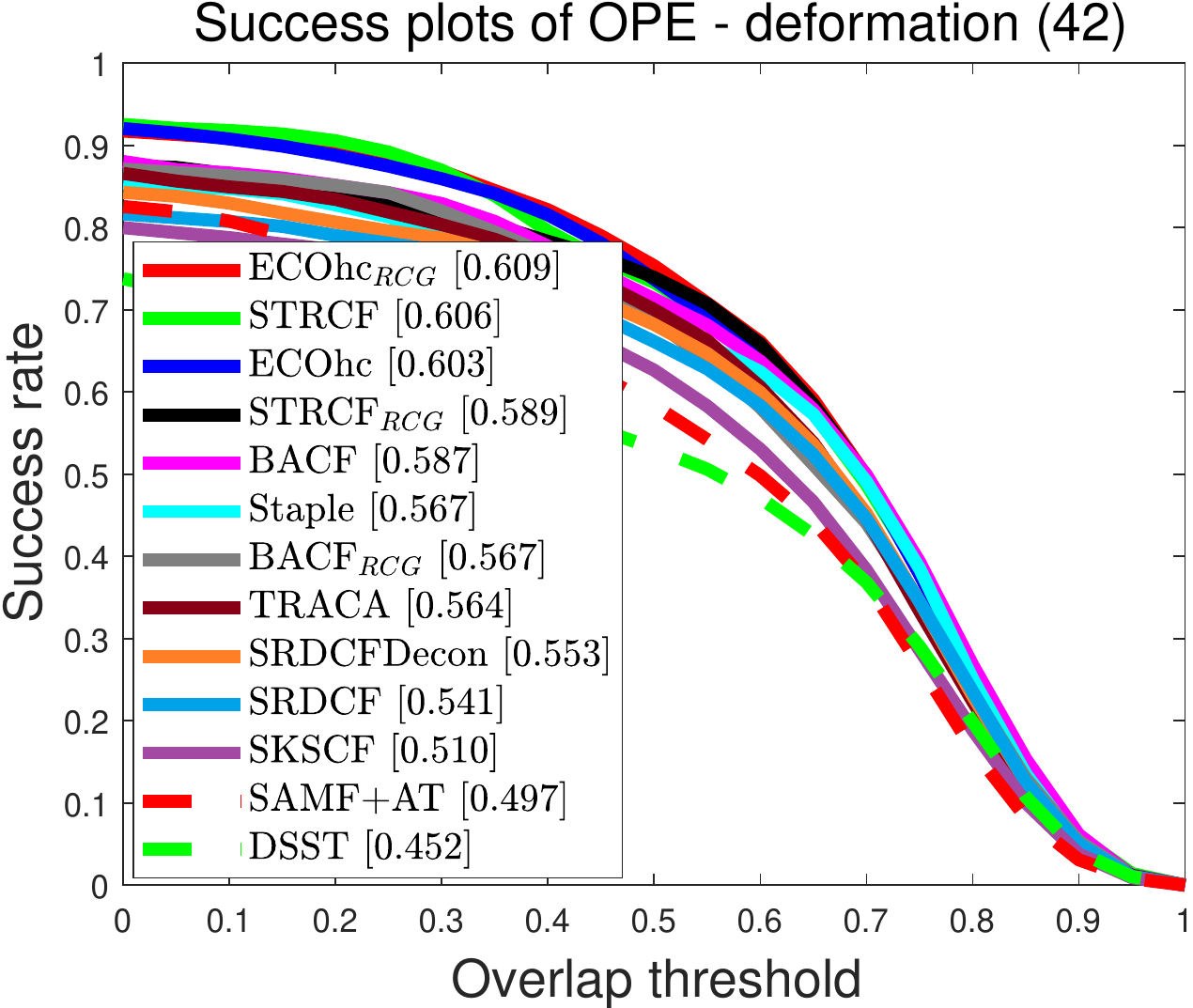}}
      \subfloat[]{\label{fig:OV}
      \includegraphics[width=0.243\textwidth]{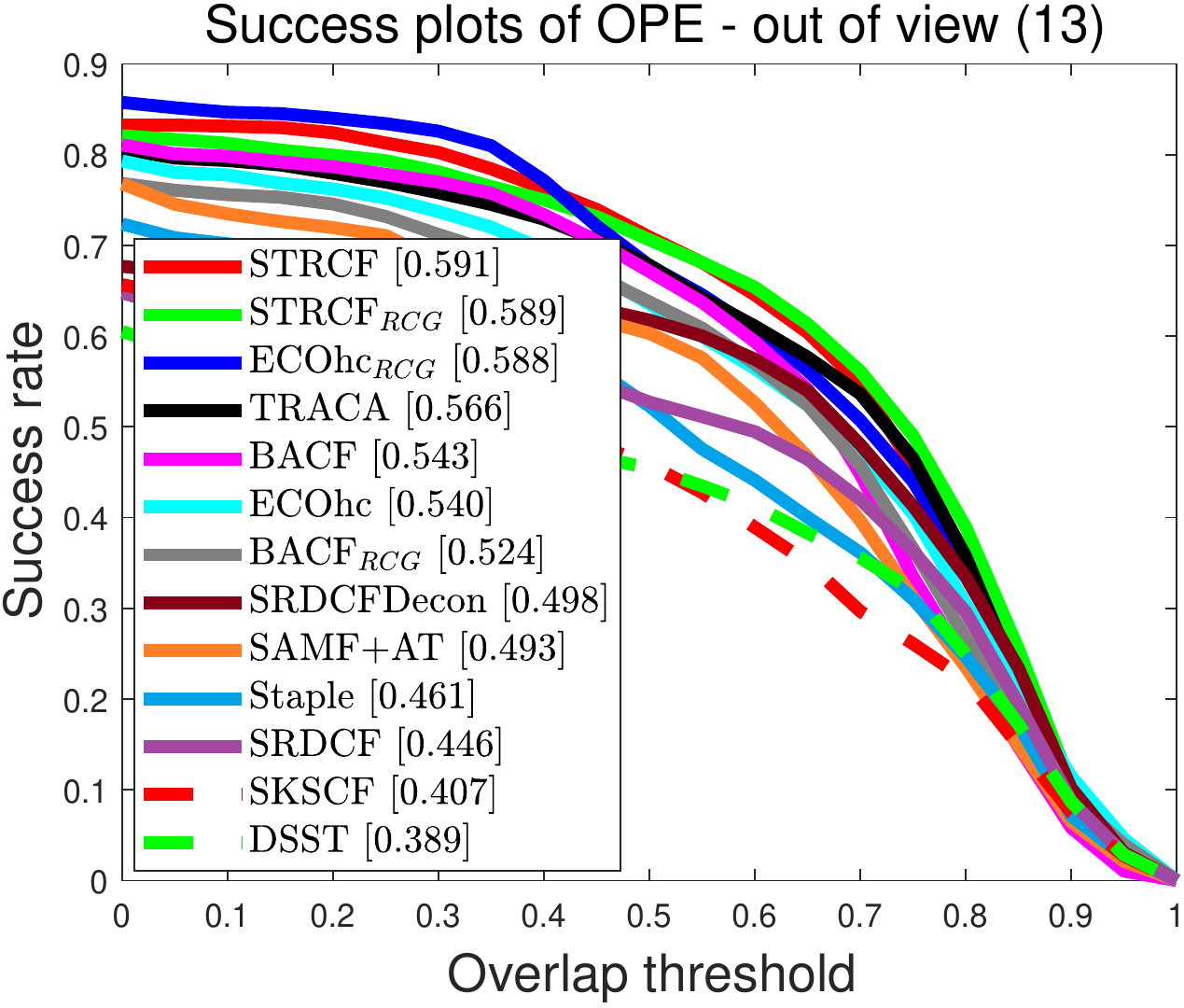}}
      \subfloat[]{\label{fig:LR}
      \includegraphics[width=0.243\textwidth]{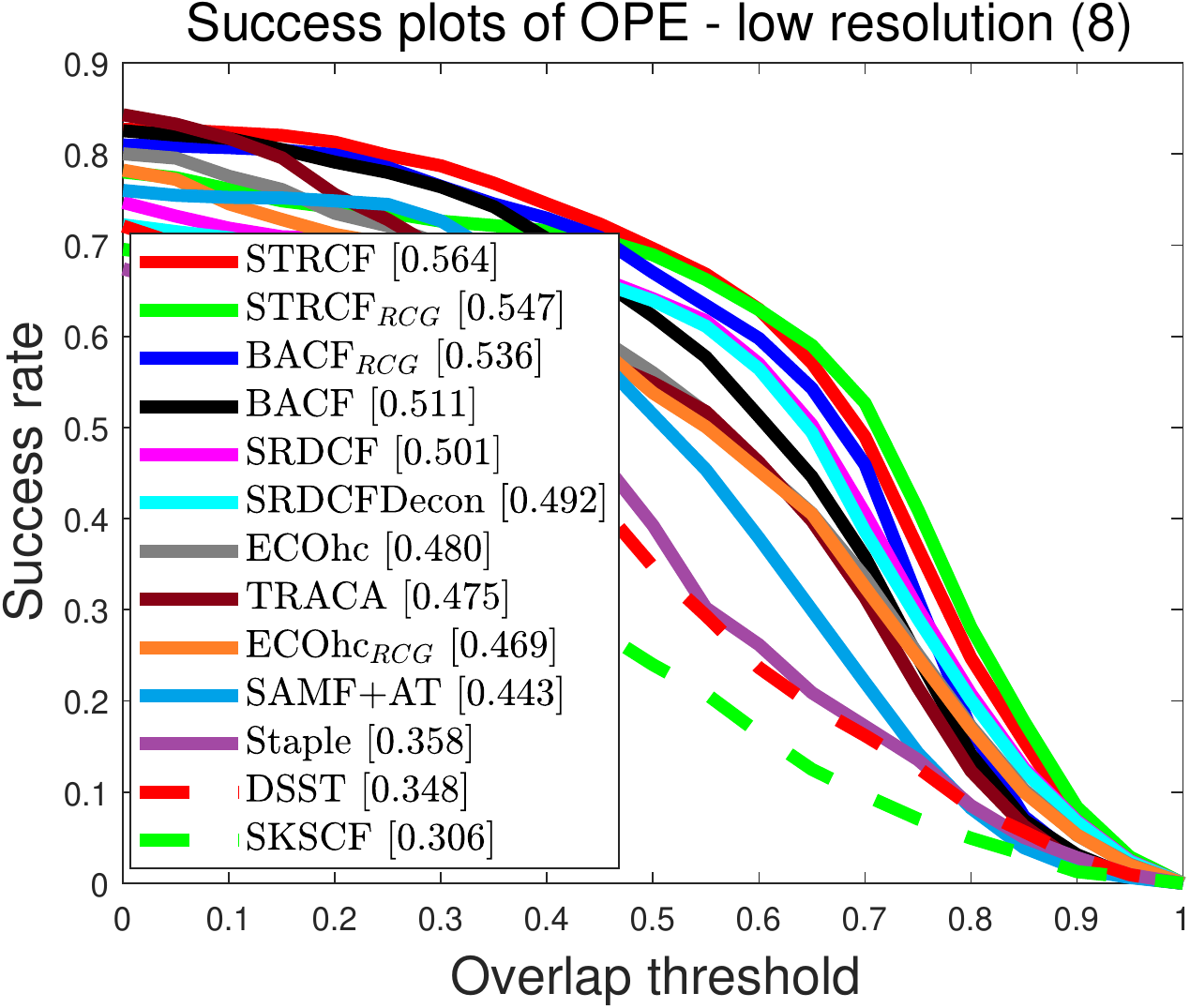}}
      \caption{Overlap success plots of the competing trackers using handcrafted features under all eleven attributes on the OTB-2015 dataset. Our methods achieve the best performance on most attributes.}
      \label{fig:attributes}
      \end{figure*}

   Using the handcrafted features, we further investigate the performance of our methods on all 11 video attributes.
   Table~\ref{Tab:OTBAttrb} gives the mean OP results of all the trackers.
   One can see that our ECO$_{RCG}$ and STRCF$_{RCG}$ obtain the rank-1 performance on 9 of all 11 video attributes.
   For the attributes \emph{motion blur, background clutter, illumination variation and occlusion}, significant improvement can be achieved by our methods.
   By removing cosine window and incorporating mask function, our methods are more effective in exploiting negative samples for model learning, and benefit the robustness of tracking performance.
   This may explain the better results of our methods when the target suffers from rapid appearance changes (e.g., motion blur, occlusion, and illumination variation) and background clutter.
   In addition, Fig.~\ref{fig:attributes} provides the AUC plots of all competing trackers using handcrafted features on all video attributes.
   It can be seen that our ECO$_{RCG}$ and STRCF$_{RCG}$ also perform favorably against the state-of-the-art methods on most attributes.

   \subsubsection{Running time}

\begin{figure}[!htbp]
      \centering
      \includegraphics[width=0.35\textwidth]{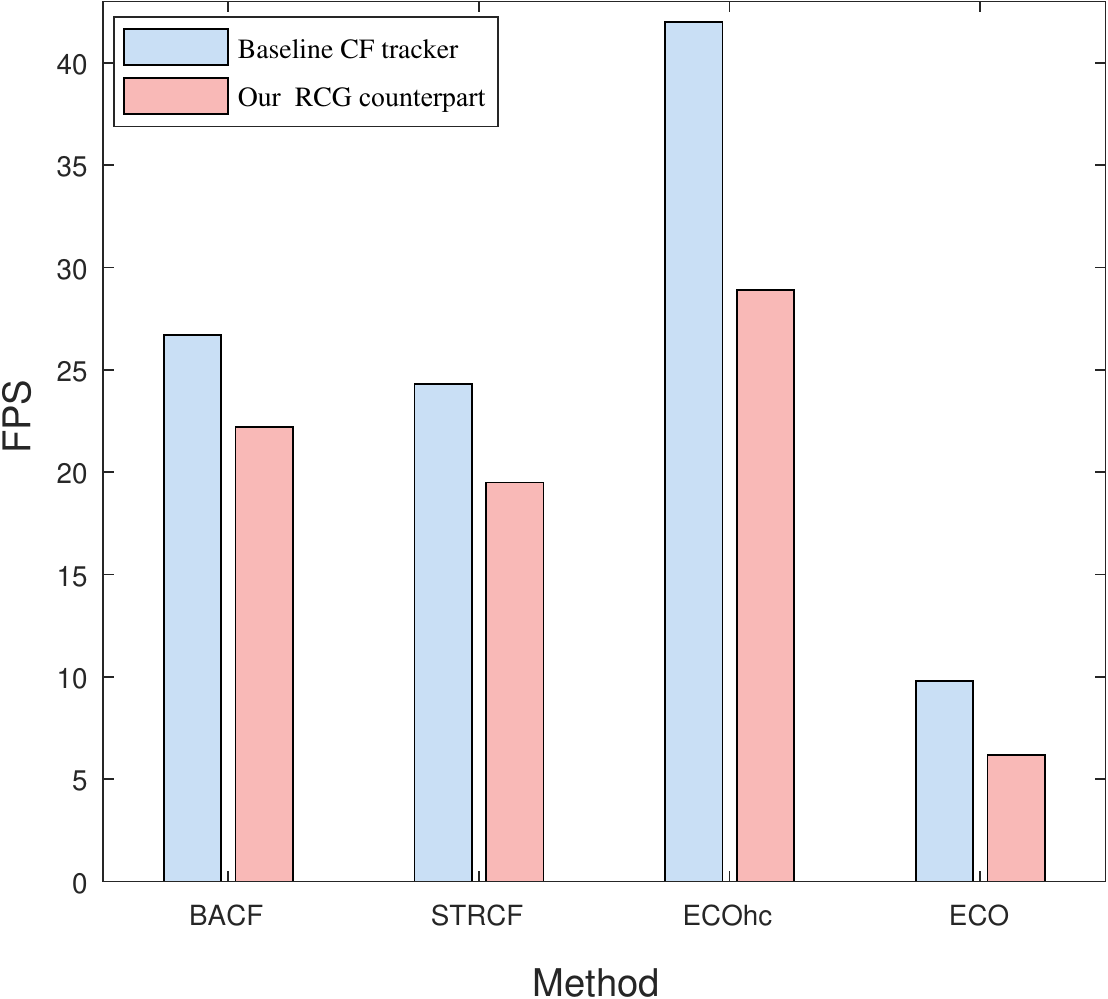}
      \caption{The tracking speed of four baseline CF trackers, i.e., BACF, STRCF, ECOhc and ECO, and their corresponding $\textbf{RCG}$ methods on OTB-2015.}
      \label{fig:speed}
   \end{figure}

   Fig.~\ref{fig:speed} reports the tracking speed (FPS) of the four baseline trackers, i.e., BACF, STRCF, ECOhc and ECO, and their corresponding $\textbf{RCG}$ methods on OTB-2015.
   It can be seen that BACF$_{RCG}$, STRCF$_{RCG}$ achieve a tracking speed of 22.2 and 19.5 FPS, moderately slower than their CF counterparts BACF (26.7 FPS) and STRCF (24.3 FPS). respectively.
   Thus, while the introduction of mask function increases the model complexity,
   the two trackers can still be efficiently solved with the ADMM algorithms, and each subproblem has its closed-form solution.
   As for the trackers with multiple base images, ECOhc$_{RCG}$ runs at approximately 70\% speed of the baseline ECOhc (42 FPS), but still maintains real-time tracking performance with 28.9 FPS.
   When extended to deep CNN features, ECO$_{RCG}$ (5.9 FPS) can run at approximately 70\% speed of its baseline ECO method (9.8 FPS).
   %

   \subsubsection{Qualitative evaluation}

\begin{figure*}[!htbp]
      \centering
      \includegraphics[width=1\textwidth]{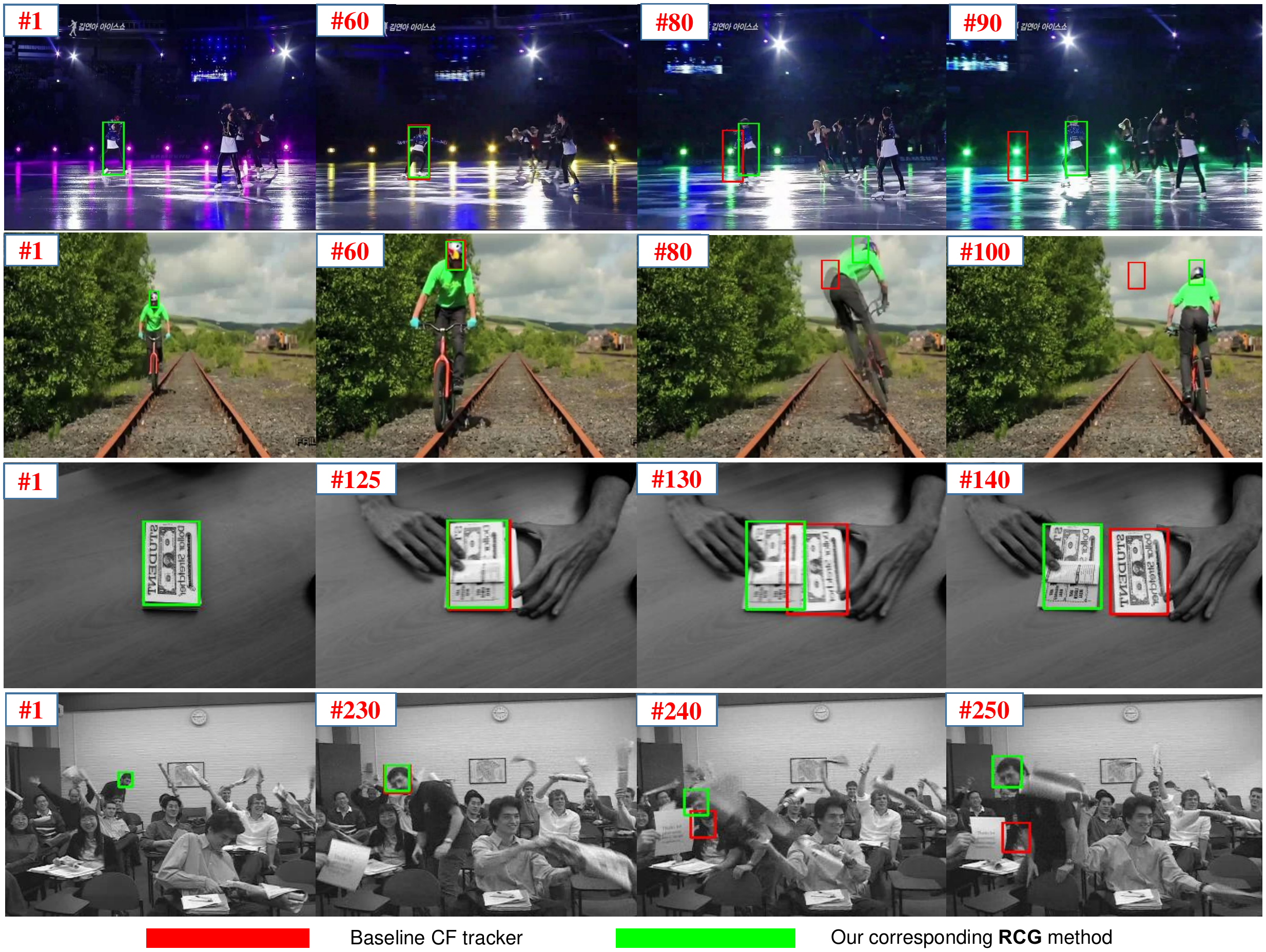}
      \caption{Visualization of tracking results on four videos for comparing four baseline CF trackers with our corresponding \textbf{RCG} methods by removing cosine window and incorporating Gaussian shaped mask function.
      The videos from top to bottom are \emph{skating1}, \emph{biker}, \emph{coupon} and \emph{freeman4}. In each row, a baseline CF tracker and our corresponding \textbf{RCG} method are applied (from top to bottom: BACF~\cite{Galoogahi2017Learning}, ECOhc~\cite{Danelljan2016ECO}, STRCF~\cite{li2018learning} and ECO~\cite{Danelljan2016ECO}).}
      \label{fig:visualization}
   \end{figure*}

   Fig.~\ref{fig:visualization} shows the qualitative results of four baseline CF trackers, i.e., BACF, STRCF, ECOhc and ECO, as well as their \textbf{RCG} counterparts.
   It can be seen from the first row that the target suffers from background clutter and illumination variation.
   In comparison to baseline BACF, BACF$_{RCG}$ can take the benefit of removing cosine window,
   thus is able to exploit more useful and uncontaminated training samples for robust model learning, thereby significantly alleviating the tracker drift issue.
   In the second row, due to the effect of motion blur, fast motion and occlusion challenges, ECOhc cannot track the target throughout the whole sequence while ECOhc$_{RCG}$ still performs well.
   %
   %

   In the last two rows, similar phenomena can also be observed when the STRCF and ECO trackers are applied to \emph{coupon} and \emph{freeman4} videos, respectively.
   In all these videos, the \textbf{RCG} method consistently outperforms its baseline CF counterpart, indicating the effectiveness of removing cosine window and incorporating with mask function.
   %
   %

   \begin{figure}[!htbp]
      \centering
      \subfloat[]{\label{fig:TempleAUCfig(1)}
         \includegraphics[width=0.24\textwidth]{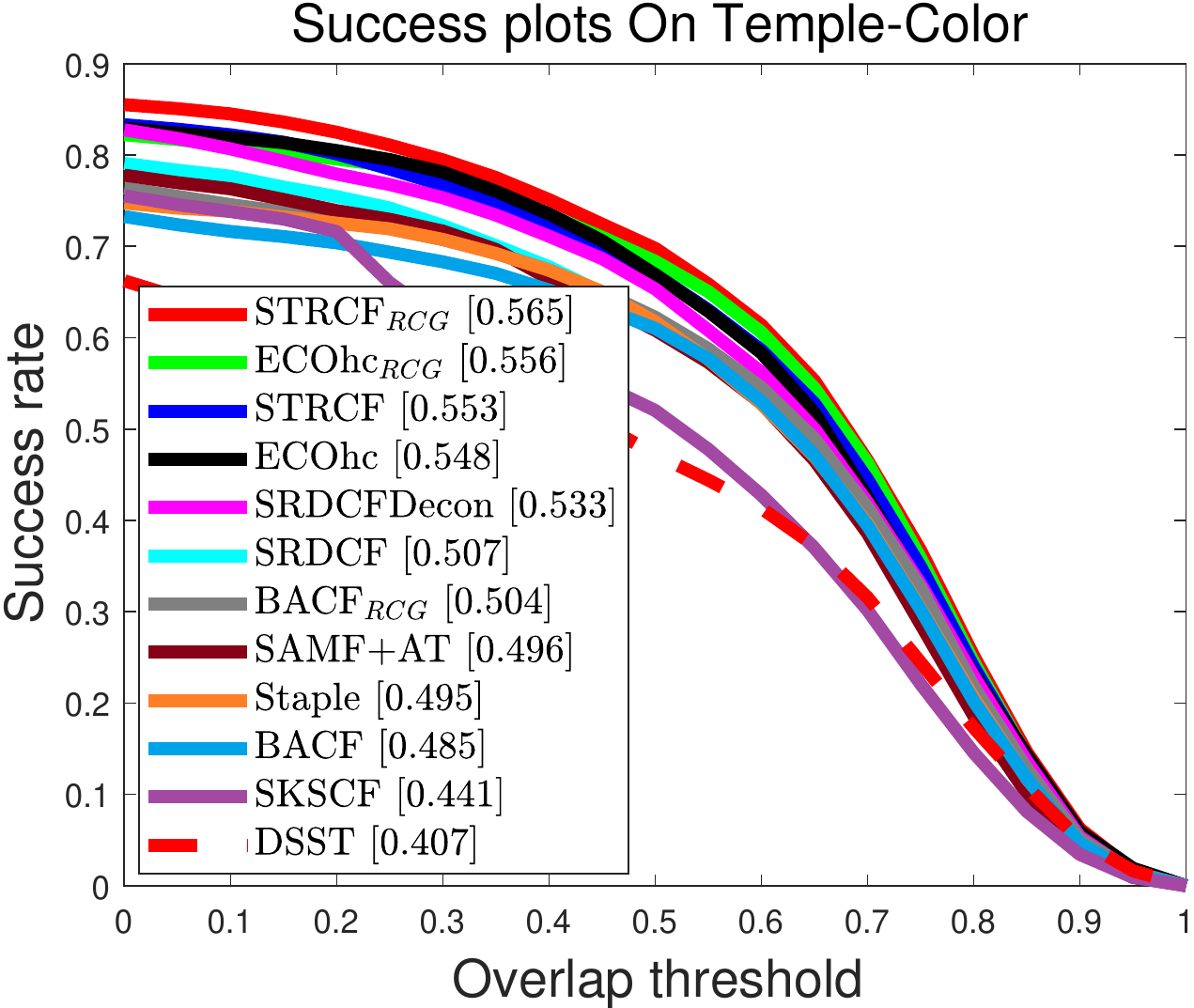}}
      \subfloat[]{\label{fig:TempleAUCfig(2)}
         \includegraphics[width=0.24\textwidth]{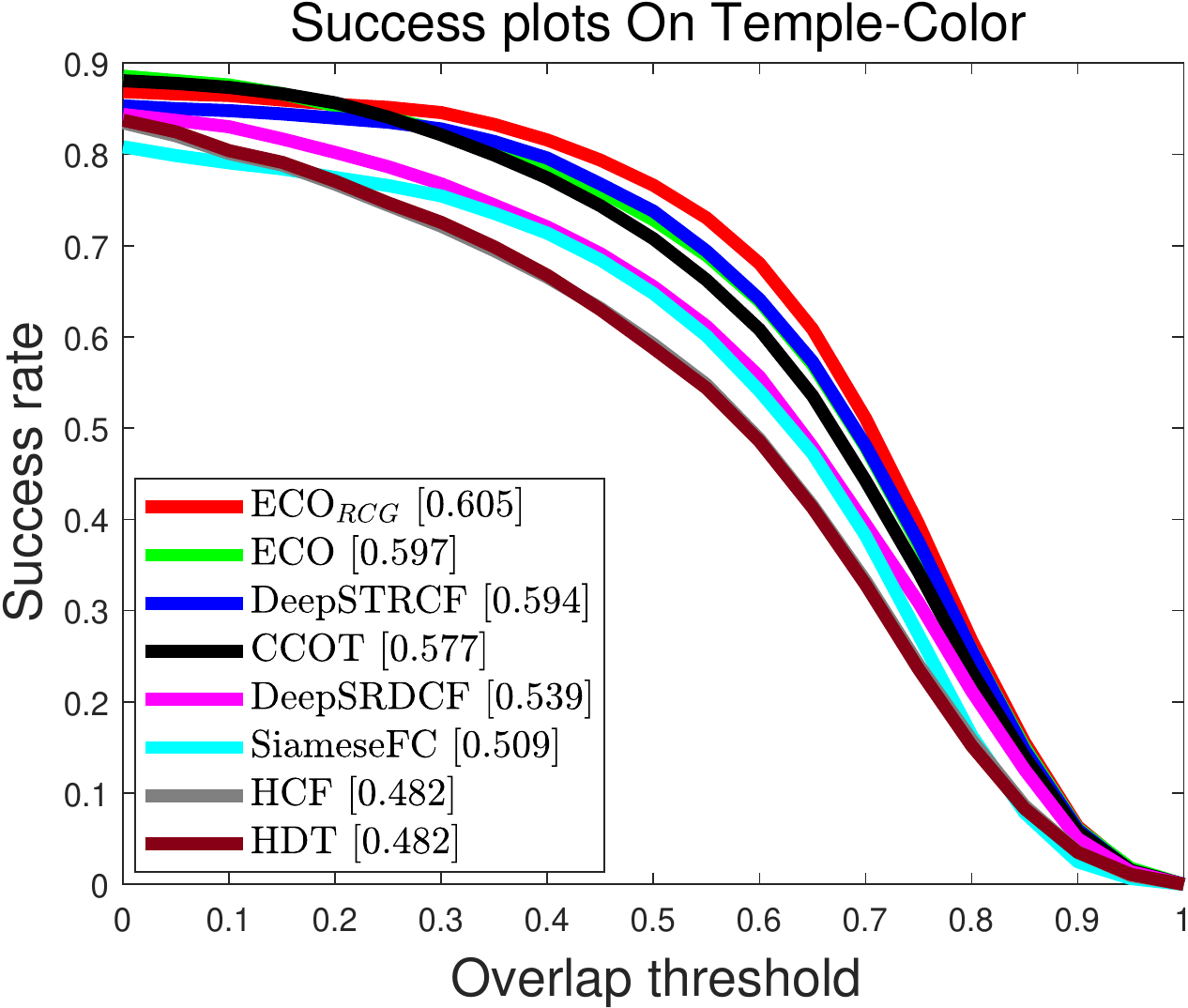}}
      \caption{Comparison of overlap success plots with the state-of-the-art trackers on the Temple-Color dataset: (a) trackers using handcrafted features, and (b) trackers using deep CNN features.}
      \label{fig:Temple}
      \end{figure}

   \subsection{Temple-Color dataset}

   To further evaluate our methods, comparative experiments are also conducted on the Temple-Color dataset containing 129 color video sequences in total.
   Fig.~\ref{fig:Temple} shows the overlap success plots of the competing trackers using handcrafted and CNN features.
   It can be seen from Fig.~\ref{fig:Temple}(a) that our methods generally consistently improve the baseline CF trackers using handcrafted features.
   In particular, BACF$_{RCG}$, ECOhc$_{RCG}$ and STRCF$_{RCG}$ respectively outperform the baseline CF counterparts with AUC score gains of 2.4\%, 0.8\% and 1.3\%.
   Moreover, as shown in Fig.~\ref{fig:Temple}(b), ECO$_{RCG}$ also performs better than ECO by 0.8\% when using deep CNN features, further demonstrating the effectiveness of removing cosine window from CF trackers with spatial regularization.

   \section{Conclusion}\label{sec:conclusion}
   In this paper, we investigated the problems of when and how to remove cosine window from CF trackers.
   %
   %
   Our empirical analysis showed that both spatial regularization and cosine window can be utilized to alleviate boundary discontinuity.
   However, cosine window may give rise to sample contamination, while for spatial regularization a small percentage of negative samples still suffer from boundary discontinuity.
   To remove cosine window from CF trackers with spatial regularization, we introduced a binary mask function to exclude the negative samples suffering from boundary discontinuity during training.
   Furthermore, another Gaussian shaped mask function was also introduced to downweight the negative samples far from target center.
   Then, optimization algorithms were respectively developed for removing cosine window from CF trackers with single and multiple base images.


   %

   Our experiments on OTB-2015, Temple-Color and VOT-2018 showed that our methods are effective in circumventing boundary discontinuity and sample contamination, and bring moderate performance gains over their CF counterparts with cosine window.
   Our methods also perform favorably against the state-of-the-art trackers using handcrafted and deep CNN features.
   %


\bibliographystyle{IEEEtran}
\bibliography{egbib}




\end{document}